\documentclass[10pt,journal,compsoc]{IEEEtran}
\ifCLASSOPTIONcompsoc
  \usepackage[nocompress]{cite}
\else
  \usepackage{cite}
\fi
\ifCLASSINFOpdf
\else
\fi
\usepackage{bbding}

\usepackage{amsfonts}

\usepackage{amsmath}

\usepackage{graphicx}

\usepackage{float}  

\usepackage{subfigure}

\usepackage{bm}

\usepackage{courier}

\usepackage{booktabs}

\usepackage{multirow}

\usepackage{rotating}

\usepackage{amsthm}

\usepackage{footnote}

\usepackage{color}

\usepackage{stfloats}

\usepackage{threeparttable}

\usepackage{adjustbox}

\usepackage{amssymb}

\usepackage[table]{xcolor}

\newtheorem{definition}{Definition}

\newtheorem{property}{Property}

\newtheorem{formulation}{Formulation}

\usepackage{color}

\definecolor{mygray}{gray}{.85}
\definecolor{mygray1}{gray}{.7}
\definecolor{mygray2}{gray}{.93}

\newcommand{\tabincell}[2]{\begin{tabular}{@{}#1@{}}#2\end{tabular}}

\makeatletter
\newcommand{\thickhline}{%
	\noalign {\ifnum 0=`}\fi \hrule height 1pt
	\futurelet \reserved@a \@xhline
}
\makeatother

\begin{document}
%
\title{Hierarchical Invariance for Robust and Interpretable Vision Tasks at Larger Scales}
%
%
%
%

\author{Shuren~Qi,~Yushu~Zhang,~Chao~Wang,~Zhihua~Xia,~Xiaochun~Cao,~and~Jian~Weng%
\IEEEcompsocitemizethanks{
	\IEEEcompsocthanksitem S. Qi, Y. Zhang, and C. Wang are with the College of Computer Science and Technology, Nanjing University of Aeronautics and Astronautics, Nanjing, China (e-mail: {shurenqi, yushu, c.wang}@nuaa.edu.cn).
	\IEEEcompsocthanksitem Z. Xia and J. Weng are with the College of Cyber Security, Jinan University, Guangzhou, China (e-mail: xiazhihua@jnu.edu.cn, cryptjweng@gmail.com).
	\IEEEcompsocthanksitem X. Cao is with the School of Cyber Science and Technology, Shenzhen Campus of Sun Yat-sen University, Shenzhen, China (e-mail: caoxiaochun@mail.sysu.edu.cn).}
}

%
%

\markboth{S. Qi \MakeLowercase{\textit{et al.}}: Hierarchical Invariance for Robust and Interpretable Vision Tasks at Larger Scales} {S. Qi \MakeLowercase{\textit{et al.}}: Hierarchical Invariance for Robust and Interpretable Vision Tasks at Larger Scales}
%



\IEEEtitleabstractindextext{%
\begin{abstract}
Developing robust and interpretable vision systems is a crucial step towards trustworthy artificial intelligence. In this regard, a promising paradigm considers embedding task-required invariant structures, e.g., geometric invariance, in the fundamental image representation. However, such invariant representations typically exhibit limited discriminability, limiting their applications in larger-scale trustworthy vision tasks. For this open problem, we conduct a systematic investigation of \emph{hierarchical invariance}, exploring this topic from theoretical, practical, and application perspectives. At the theoretical level, we show how to construct over-complete invariants with a Convolutional Neural Networks (CNN)-like hierarchical architecture yet in a fully interpretable manner. The general blueprint, specific definitions, invariant properties, and numerical implementations are provided. At the practical level, we discuss how to customize this theoretical framework into a given task. With the over-completeness, discriminative features w.r.t. the task can be adaptively formed in a Neural Architecture Search (NAS)-like manner. We demonstrate the above arguments with accuracy, invariance, and efficiency results on texture, digit, and parasite classification experiments. Furthermore, at the application level, our representations are explored in real-world forensics tasks on adversarial perturbations and Artificial Intelligence Generated Content (AIGC). Such applications reveal that the proposed strategy not only realizes the theoretically promised invariance, but also exhibits competitive discriminability even in the era of deep learning. For robust and interpretable vision tasks at larger scales, hierarchical invariant representation can be considered as an effective alternative to traditional CNN and invariants.
\end{abstract}

}

\maketitle

\IEEEdisplaynontitleabstractindextext

%
\IEEEpeerreviewmaketitle

\IEEEraisesectionheading{\section{Introduction}\label{sec:intro}}

%
%
%
%

\IEEEPARstart{T}{he} trustworthiness is a rising topic in modern Artificial Intelligence (AI) communities \cite{ref1}. Over the past decade, deep learning techniques, especially Convolutional Neural Networks (CNN), have led to breakthrough results in numerous AI tasks, e.g., processing human perceptual information \cite{ref2}, playing board games \cite{ref3}, and solving hard science problems \cite{ref4}. More recently, their applications are expanding into trust-related scenarios, e.g., biometrics \cite{ref5}, medical diagnostics \cite{ref6}, self-driving cars \cite{ref7}, and misinformation detection \cite{ref8}. In such scenarios, the robustness and interpretability of AI systems are crucial \cite{ref9}: 1) \emph{robustness} means the performance of system is stable for intra-class variations on the input; 2) \emph{interpretability} means the behavior of system can be understood or predicted by humans.

Integrating invariant structures into image representations is a principled design towards robust and interpretable vision systems \cite{ref10}. Specifically, representations play a fundamental role in visual systems, where the system is generally built on meaningful representations of digital images (rather than the raw data) \cite{ref11}. Note that the proper design/learning of such representations in fact relies on priors w.r.t. the task of interest \cite{ref12}. Here, the concept of \emph{symmetry} from the Erlangen Program \cite{ref13, ref14} may be the most fruitful prior -- informally, a symmetry of a system is a transformation that leaves a certain property of system invariant. Symmetry priors are ubiquitous in vision tasks, e.g., translation as a symmetry of the object classification system where object category is invariant under translation \cite{ref15}.

Next, we begin by providing some historical perspectives on the invariance in the development of image representations. The quest for invariance is as old as the ﬁeld of computer vision itself, spanning both hand-crafted and learning approaches \cite{ref16}:

\begin{itemize}
	\item In the hand-crafted approach, symmetry priors (e.g., invariance and equivariance) w.r.t. geometric transformations (e.g., translation, rotation, and scaling) have been recognized as main ideas in designing representations. Such ideas cover almost all classical and state-of-the-art methods, from global features (e.g., moment invariants \cite{ref17}), to local sparse features (e.g., SIFT \cite{ref18}), and to local dense features (e.g., DAISY \cite{ref19}). However, these hand-crafted representations are all fixed in design, relying on \emph{(under)-complete} dictionaries, and therefore fail to provide sufficient discriminability at larger scales, e.g., ImageNet classification task \cite{ref20}.
	\item In the learning approach, the CNN achieves \emph{over-complete} representations of strong discriminative power for larger-scale vision tasks, through a cascade of learnable nonlinear transformations. As a textbook view of deep learning, representations should be learned not designed \cite{ref2}. Therefore, classical CNN representations are equipped with very few symmetry priors, typically just translation equivariance \cite{ref15}, but has recently been proven to no longer hold in deeper layers of the CNN with downsampling structures \cite{ref21}. In general, these learning representations lack robustness and interpretability guarantees, e.g., the presence and understanding of adversarial perturbations \cite{ref22}, and therefore cannot be well extended to trustworthy tasks \cite{ref23}.
\end{itemize}

Historically, to a certain extent, efforts at invariance and discriminability have developed independently in hand-crafted and learning approaches. The compatibility between invariance and discriminability has emerged as a tricky problem when moving towards trustworthy AI.

\begin{table*}[t]
	\renewcommand{\arraystretch}{1.3}
	\caption{Conceptual Comparison with Related Research Approaches.}
	\centering
	\begin{tabular}{cccccc}
		\toprule
		Approach & Traditional invariance & Traditional CNN & Scattering networks &  Equivariant networks  & Hierarchical invariance\\
		\midrule
		\rowcolor{mygray2}Discriminative &   & \checkmark & \checkmark  & \checkmark & \checkmark \\
		Robust &  \checkmark &      & \checkmark     & \checkmark     & \checkmark \\
		\rowcolor{mygray2}Interpretable & \checkmark &      & \checkmark     & \checkmark     & \checkmark \\
		Efficient &  \checkmark &  &     &     & \checkmark \\
		\bottomrule
	\end{tabular}%
\end{table*}

\subsection{State of the Art and Motivation}

In trust-related scenarios at larger scales, recent advances are seeking more advanced invariant designs of image representations, fulfilling discriminability, robustness, and interpretability simultaneously.

\begin{itemize}
	\item On the discriminability of hand-crafted representations, researchers introduce successful experiences behind learning representations, especially cascading and over-complete designs. The most representative work is Invariant Scattering Convolution Networks \cite{ref24}, where the classical wavelet transform is expanded into an over-complete representation with deep cascading. Unlike typical CNN: 1) regarding the architecture, convolutional layers are defined by fixed wavelet ﬁlters, with modulus-based nonlinearity, but without subsequent pooling; 2) regarding the representation properties, the architecture yields the translation equivariance and certain robustness w.r.t. non-linear deformations. Following theoretical works further explored various geometric invariants \cite{ref25}, more general mathematical formulations \cite{ref26}, and the potential for improving the efficiency, interpretability, and robustness of state-of-the-art CNN techniques \cite{ref27}. Regarding applications, they provided competitive results in a variety of tasks on audio \cite{ref28}, image \cite{ref24, ref25, ref27} and graph \cite{ref29} data, some of which are even interdisciplinary \cite{ref30, ref31}. With similar design goals and paths, we consider Scattering Networks as a main competitor for our work.
	\item On the robustness and interpretability of learning representations, researchers introduce successful experiences behind hand-crafted representations, especially invariance and equivariance designs. The most representative work is Group Equivariant Convolutional Networks \cite{ref32}, where the classical convolution is generalized to a new definition on the symmetry group. Unlike typical CNN: 1) regarding the architecture, convolutional layers are learned but with new structure inspired by symmetry priors, e.g., re-parameterizing the filter to control symmetry; 2) regarding the representation properties, the architecture commonly provides the joint equivariance for translation and rotation. Following theoretical works further explored the equivariance for rotation \cite{ref33, ref34, ref35}, flipping \cite{ref33}, scaling \cite{ref36, ref37}, and their combination \cite{ref38} from various mathematical theories, including steerable ﬁlters \cite{ref33}, harmonic analysis \cite{ref34}, scale space \cite{ref37}, Lie groups \cite{ref39}, and B-spline interpolation \cite{ref40}. Regarding applications, they played a key role in low-level vision tasks \cite{ref41}, especially scientific discoveries with symmetry priors \cite{ref42, ref43}. With similar design goals but on the learning path, we consider Equivariant Networks as a secondary competitor for our work.
\end{itemize}

\emph{Motivation.} Despite starting from different theories, the above state-of-the-art methods exhibit common problems in the implementation efficiency and representation capability. An over-simpliﬁed interpretation on the technical level of such methods is as follows \cite{ref38}: The input feature map is convolved with symmetry versions of the same filter to obtain multi-channel features, where the distortion on the input (e.g., rotation) corresponds to the cyclic shift between channels, and hence the invariance is achieved by pooling across channels. 1) Regarding the implementation efficiency, this parallel framework leads to an exponential expansion on the computational size (w.r.t. the sampling rate on symmetry), especially for learning representations where the introduced new learnable parameters make the training more challenging. 2) Regarding the representation capability, the discrete sampling on the symmetry of filters raises a tricky trade-off between invariance and discriminability -- higher sampling rate implies better invariance, but the resulting computational cost restricts the overall size of representation networks (going deeper or wider), and hence the improvement of discriminability.

\subsection{Contributions}

As a potential step towards solving the above open problem, we conduct a systematic investigation of \emph{hierarchical invariance}, exploring this topic from theoretical, practical, and application perspectives.

As summarized in Table 1, our approach stands nicely between the two extremes, i.e., traditional invariance \cite{ref17} and CNN \cite{ref21} w.r.t. discriminability, robustness, and interpretability. Compared with the recent scattering \cite{ref24} and equivariant \cite{ref32} networks, our approach is characterized by a more efficient design. Here, the equivariance is continuous/one-shot, holds across layers, eliminating the need for complex symmetry sampling, parallel framework, and cross-channel pooling. Therefore, it exhibits better efficiency, and also allowing more flexibility in enlarging the network size (going deeper or wider) to increase the representation capacity.

Our main contributions are as follows:

\emph{Theory.} We propose a new framework for robust and interpretable image representation, named Hierarchical Invariant Representation (HIR), by extending the classical theory of moment invariants to cascade transformations. Starting from an ideal blueprint for hierarchical invariance, we formalize the over-complete moment invariants by an efficient hierarchical structure, with better trade-off between invariance and discriminability than traditional invariants and CNN. Note that the HIR exhibits continuous and one-shot equivariance w.r.t. translations, rotations, and flips at each intermediate layer, which is not available in current state-of-the-art algorithms. We also provide some fast and accurate numerical implementations of HIR, which are generic for arbitrary basis functions.

\emph{Practice.} We explore the practical flexibility of this theoretical framework, covering a wide range of potential designs to better match a given vision task. The theory is specified into a class of networks, involving details about the topologies, layers, and parameters. In particular, we define a concept of frequency pooling to satisfy the common requirements from interpretability, invariance, and discriminability. For empowering the data adaptability of hand-crafted HIR, we also provide an architecture searching strategy based on the over-completeness.

\emph{Application.} We validate the effectiveness of HIR in various simulation experiments and real-world applications. Pattern classification experiments are performed on typical sets of texture, digit, and parasite images, exhibiting state-of-the-art accuracy, invariance, and efficiency under diverse task scales and geometric variants. The direct applications to real-world forensics, i.e., detections of adversarial perturbations \cite{ref44} and Artificial Intelligence Generated Content (AIGC) \cite{ref45}, also demonstrate the competitive discriminability even in the era of deep learning.

\section{Foundations}

As mentioned earlier, this work develops from the theory of moment invariants. Therefore, we begin with a brief review on the foundations of moment invariants, covering some concepts, notations, and definitions from our previous works.

\subsection{Global and Local Representations}

In general, classical moments and moment invariants are \emph{global} representations of images, where the theory is built on the following definition \cite{ref17}:
\begin{equation}
	\left< f,{V_{nm}} \right>  = \iint_{D}{{V_{nm}^*(x,y)f(x,y)dxdy}},
\end{equation}
where $f$ is the image function, ${V_{nm}}$ is the basis function with order parameter $(n,m) \in \mathbb{Z}^2$ on domain $D$, and $*$ is the complex conjugate. Note that the domains of $f$ and ${V_{nm}}$ in (1) have the same/similar location and scale, implying the global nature of the representation information.

With the sparse prior and geometric prior for natural images, two typical constraints, i.e., orthogonality and rotation invariance, often imposed on the explicit definition of ${V_{nm}}$, leading to the following polar form:
\begin{equation}
	\left< f,{V_{nm}} \right>  = \iint_{D}{{R_n^*(r)A_m^*(\theta )f(r,\theta )rdrd\theta }},
\end{equation}
where ${V_{nm}}(\underbrace {r\cos \theta }_x,\underbrace {r\sin \theta }_y) \equiv {V_{nm}}(r,\theta )$ is separated as the product of the angular basis function ${A_m}(\theta ) = \exp (\bm{j}m\theta )$ ($\bm{j} = \sqrt { - 1} $) and the radial basis function ${R_n}$, subject to the weighted orthogonality condition $\int\limits_0^1 {{R_n}(r)R_{n'}^*(r)rdr}  = \frac{1}{{2\pi }}{\delta _{nn'}}$. Note that the basis function ${V_{nm}} = {R_n}{A_m}$ in (2) is orthogonal on $D$, and the magnitude of $\left< f,{V_{nm}} \right>$ is invariant to the rotation on the image $f$ (see \cite{ref17} for a survey).

In our recent work, moments and moment invariants are extended to \emph{local} representations of images, where the theory is built on the following definition \cite{ref46}:
\begin{equation}
	\left< f,V_{nm}^{uvw} \right>  = \iint_{D}{{R_n^*(r')A_m^*(\theta ')f(x,y)dxdy}},
\end{equation}
where the new basis function $V_{nm}^{uvw}$ introduces position parameter $(u,v)$ and scale parameter $w$. It can be interpreted as a translated and scaled version of the global ${V_{nm}}$ with the following coordinate relationship:
\begin{equation}
	\left\{ {\begin{array}{*{20}{l}}
			{r' = \frac{1}{w}\sqrt {{{(x - u)}^2} + {{(y - v)}^2}} }\\
			{\theta ' = \arctan (\frac{{y - v}}{{x - u}})}
	\end{array}} \right.,
\end{equation}
where the domain is a disk centered at $(u,v)$ and with radius $w$: $D = \{ (x,y):{(x - u)^2} + {(y - v)^2} \le {w^2}\}$. Note that (3) allows the domain of $V_{nm}^{uvw}$ to be built in different positions and scales w.r.t. the domain of $f$, implying the local nature of the representation information. Also, the classical definition (2) is in fact a special case of the new definition (3) with $(u,v) = (0,0)$ and $w = 1$ (see \cite{ref46} for details).

\subsection{Invariance, Equivariance, and Covariance}

The terms of invariance, equivariance, and covariance appear in the fields of computer vision, graphics, geometry, and physics, with similar but slightly different definitions. In this paper, we use the following identities to generally denote such terms \cite{ref47, ref48}:
\begin{itemize}
	\item invariance – ${\cal R}({\cal D}(f)) \equiv {\cal R}(f)$,
	\item equivariance – ${\cal R}({\cal D}(f)) \equiv {\cal D}({\cal R}(f))$,
	\item covariance – ${\cal R}({\cal D}(f)) \equiv {\cal D}'({\cal R}(f))$,
\end{itemize}
where ${\cal R}$ is a representation, ${\cal D}$ is a degradation, and ${\cal D}'$ is a composite function of ${\cal D}$. Note that invariance and equivariance are special cases of covariance with ${\cal D}' = {\rm{id}}$ and ${\cal D}' = {\cal D}$.

Starting from the above terms and the local representation (3), one can verify that  $\left< f,V_{nm}^{uvw} \right>$ exhibits the following properties w.r.t. translation, rotation, flipping, and scaling on images (see \cite{ref46} for details).

The image \emph{translation} leads to
\begin{equation}
	\begin{split}
	&\left< f(x + \Delta x,y + \Delta y),V_{nm}^{uvw}(x,y) \right> \\
	&=  \left< f(x,y),V_{nm}^{(u + \Delta x)(v + \Delta y)w}(x,y) \right> ,
	\end{split}
\end{equation}
where $(\Delta x,\Delta y)$ is the translation offset of the image $f$. Note that the same $(\Delta x,\Delta y)$ appears in position parameter $(u,v)$, implying the equivariance w.r.t. the image translation.

Since the translation equivariance holds, the following analysis (6) $\sim$ (8) will consider only center-aligned geometric transformations, i.e., we can restrict $(u,v) = (0,0)$ without loss of generality.

\begin{figure*}[t]
	\centering
	\includegraphics[scale=0.45]{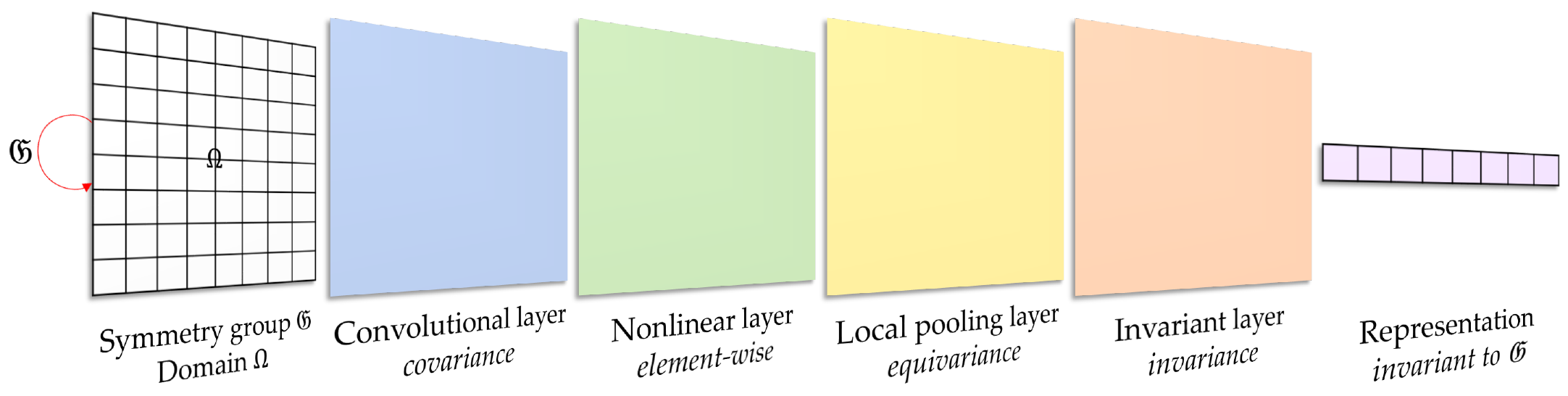}
	\centering
	\caption{The blueprint of hierarchical invariance, where the image information is able to pass through each intermediate layer in a geometrically controllable manner, and on the last layer, the invariant features are allowed by compact designs, with also sufficient information.}
\end{figure*}

The image \emph{rotation} leads to
\begin{equation}
	\begin{split}
		&\left<f(r,\theta  + \phi ),V_{nm}^{uvw}(r',\theta ') \right> \\
		&=  \left<f(r,\theta ),V_{nm}^{uvw}(r',\theta ') \right> A_m^*( - \phi ),
	\end{split}
\end{equation}
with $(u,v) = (0,0)$, where $\phi$ is the rotation angle w.r.t. the center of the image $f$. Note that the same $\phi$ appears in phase of the representation, implying the covariance w.r.t. the center-aligned rotation. It is straightforward that the covariance (6) will specialize to the invariance when taking the magnitude as $| \left< f(r,\theta  + \phi ),V_{nm}^{uvw}(r',\theta ') \right> | = | \left< f(r,\theta ),V_{nm}^{uvw}(r',\theta ') \right> |$.

The image \emph{flipping} leads to
\begin{equation}
	\begin{split}
		\left<f(r, - \theta ),V_{nm}^{uvw}(r',\theta ')\right>
		= (\left<f(r,\theta ),V_{nm}^{uvw}(r',\theta ')\right>)^*,
	\end{split}
\end{equation}
with $(u,v) = (0,0)$, where $f(r, - \theta )$ is a vertically flipped version of the image $f$ w.r.t. the center. Note that center-aligned vertical flipping again only affects the phase of the representation, implying the covariance similar to (6). As for other flipping orientations, the same conclusion can be derived from the composite of rotation and vertical flipping. It is straightforward that the joint invariance of center-aligned rotation and flipping is hold when taking the magnitude of the representation.

The image \emph{scaling} leads to
\begin{equation}
	\begin{split}
		\left<f(sx,sy),V_{nm}^{uvw}(x,y)\right>
		= \left<f(x,y),V_{nm}^{uv(ws)}(x,y)\right>,
	\end{split}
\end{equation}
with $(u,v) = (0,0)$, where $s$ is the scaling factor w.r.t. the center of the image $f$. Note that the same $s$ appears in scale parameter $w$, implying the covariance w.r.t. center-aligned scaling.

For the representation properties when $(u,v) \ne (0,0)$, they can be derived from the composite of translation with center-aligned rotation, flipping, and scaling, respectively. Hence, the magnitude of the representation has \emph{joint equivariance for any translation, rotation, and flipping} on $(u,v)$ domain, as well as \emph{covariance for any scaling} on $w$ domain.

\section{Hierarchical Invariance: Theory}

This section focuses on the theoretical aspects of the proposed HIR.

As a high-level intuition, we formalize the blueprint of hierarchical invariance, rethinking the typical modules of CNN representations. Starting from the invariant theory, we then define new modules with their compositions to fulfill such blueprint, along with representation property justifications and efficient numerical implementations. For a summary of this theoretical section, we discuss the criticisms and developments of the proposed idea versus typically concepts.

\subsection{Blueprint of Hierarchical Invariance}

Our goal is to achieve such a CNN-like hierarchical architecture -- the image information is able to pass through each intermediate layer in a geometrically controllable manner, and on the last layer, the invariant features are allowed by compact designs, with also sufficient information. In this paper, this ideal representation structure is termed as hierarchical invariance.

Motivated by the goal, we rethink several typical modules of CNN representation and formalize a blueprint of such modules for hierarchical invariance, as shown in Fig. 1.

\begin{formulation} \textbf{(Hierarchical invariance blueprint).}
	The set of feature maps (including input images) is denoted as $X(\Omega ,{\rm H}) \triangleq \{ M(i,j;k):\Omega  \to {\rm H}\} $ with the discrete domain $(i,j) \in \Omega $ and complex-valued channels ${\mathfrak{C}^k} \in {\rm H}$, where $\Omega '$ and ${\rm H}'$ are the variants of $\Omega $ and ${\rm H}$ respectively under certain operators, e.g., downsampling, and  $\mathfrak{G}$ is a group modeling all the symmetries of interest over the $\Omega $. We formalize the following modules for hierarchical invariance:
	\begin{itemize}
		\item The \emph{convolutional} layer $\mathbb{C}:X(\Omega ,{\rm H}) \to X(\Omega ',{\rm H}')$ captures local features by convolution operations. The geometric principle of $\mathbb{C}$ is the \emph{covariance} for the symmetry group $\mathfrak{G}$, i.e., there exists a predictable $\mathfrak{g}'$ such that $\mathbb{C}(\mathfrak{g}M) = \mathfrak{g}'\mathbb{C}(M)$ for any $\mathfrak{g} \in \mathfrak{G}$ and $M \in X$, where the covariance specializes to the equivariance when $\mathfrak{g}' = \mathfrak{g}$.
		\item The \emph{nonlinear} layer $\mathbb{S}:X(\Omega ,{\rm H}) \to X(\Omega ,{\rm H}')$ introduces the non-linearity in features for discriminative purposes, with an activation function $\sigma $ such that  $(\mathbb{S}M)(i,j) = \sigma (M(i,j))$, where the composition of convolutional and nonlinear layers, i.e., $\mathbb{S} \circ \mathbb{C}$, is also covariant for the group $\mathfrak{G}$, due to the \emph{element-wise} act of $\sigma $.
		\item The \emph{local pooling} layer $\mathbb{P}:X(\Omega ,{\rm H}) \to X(\Omega ',{\rm H})$ downsamples the plane dimensions of feature maps to reduce computational complexity, such that $\Omega ' \subseteq \Omega $. The geometric principle of $\mathbb{P}$ is the (approximately) \emph{equivariance} for any $\mathfrak{g}'$ produced by $\mathbb{C}$, i.e., $\mathbb{P}(\mathfrak{g}'M) \simeq \mathfrak{g}'\mathbb{P}(M)$, allowing the subsequent convolutional layer $\mathbb{C}$ to continue capturing such covariant features, where the composition $\mathbb{P} \circ \mathbb{S} \circ \mathbb{C}$ remains covariance for the group $\mathfrak{G}$.
		\item The \emph{invariant} layer $\mathbb{I}:X(\Omega ,{\rm H}) \to Y$ obtains the final vector representation via a certain global pooling over plane dimensions. The geometric principle of $\mathbb{I}$ is the \emph{invariance} for the symmetry group $\mathfrak{G}$, i.e., $\mathbb{I}(\mathfrak{g}M) = \mathbb{I}(M)$ for any $\mathfrak{g} \in \mathfrak{G}$ and $M \in X$.
	\end{itemize}
\end{formulation}

With this blueprint, HIR can be designed in a generic way, by the ordered cascading of such modules.

\subsection{Definition of Layer and Path}

From the invariant theory of Section 2, we will give a class of definitions for layers $\mathbb{C}$, $\mathbb{S}$, $\mathbb{P}$, and $\mathbb{I}$, satisfying the hierarchical invariance blueprint.

\begin{definition} \textbf{(Convolutional layer).}
	For the input feature map $M(i,j;k)$ with $\Omega  = \{ 1,2,...,{N_i}\}  \times \{ 1,2,...,{N_j}\} $ and ${\rm H} = {\mathfrak{C}^K}$, the convolutional layer $\mathbb{C}$ is defined channel-wise as \textbf{local covariant representations} with (3) and (4) :
	\begin{equation}
		\mathbb{C}M \triangleq \left< M,V_{nm}^{uvw} \right>  = M(i,j;k) \otimes {(H_{nm}^w(i,j))^T},
	\end{equation}
	where $\otimes$ is the convolution over the $\Omega$, $( \cdot )^T$ denotes the matrix transpose, and $H_{nm}^w$ is a convolution kernel defined as:
	\begin{equation}
		H_{nm}^w(i,j) = \{ h_{nm}^{uvw}(i,j):u,v= w, (i,j) \;\mathrm{s.t.}\; {D_{ij}} \cap D \ne \emptyset \},
	\end{equation}
	where $h_{nm}^{uvw}$ is the integral value of basis function over a valid pixel region: 
	\begin{equation}
		h_{nm}^{uvw}(i,j) =  \iint_{{{D_{ij}} \cap D}}{{{(V_{nm}^{uvw}(x,y))}^*}dxdy},
	\end{equation}
	with $(i,j)$-centered pixel region ${D_{ij}} = \{ (x,y) \in [i - \frac{{\Delta i}}{2},i + \frac{{\Delta i}}{2}] \times [j - \frac{{\Delta j}}{2},j + \frac{{\Delta j}}{2}]\} $.
\end{definition}
\emph{Remark.} In the Definition 1, the convolutional layer $\mathbb{C}$ is specified by $w$ and $(n,m)$, which control the representation scale and representation frequency of $\mathbb{C}M$, respectively. Note that the $\mathbb{C}$ defined by (9) will directly inherit the representation properties (5) $\sim$ (8), it should be regarded as a linear covariant layer on the group $\mathfrak{G}$ modeling all translation, rotation, flipping, and scaling symmetries over the $\Omega $. For convenience, we denote $\mathfrak{G} = \mathfrak{G}_1 \times \mathfrak{G}_2$, where $\mathfrak{G}_1$ is the translation/rotation/flipping symmetry group and $\mathfrak{G}_2$ is the scaling symmetry group.

\begin{definition} \textbf{(Nonlinear layer).}
	For the input feature map $M(i,j;k)$ with $\Omega  = \{ 1,2,...,{N_i}\}  \times \{ 1,2,...,{N_j}\} $ and ${\rm H} = {\mathfrak{C}^K}$, the nonlinear layer $\mathbb{S}$ is defined channel-wise as a magnitude operation:
	\begin{equation}
		\mathbb{S}M = \sigma (M(i,j)) \triangleq |M(i,j;k)|,
	\end{equation}
	where $M(i,j;k)$ is complex-valued, and (12) can be written explicitly as $\sqrt {{{({\mathop{\rm Re}\nolimits} M(i,j;k))}^2} + {{({\mathop{\rm Im}\nolimits} M(i,j;k))}^2}} $.
\end{definition}
\emph{Remark.} With the Definition 2 and Section 2.2, the composition of convolutional and nonlinear layers will exhibit the \emph{joint equivariance of translation, rotation, and flipping}, i.e., $\mathbb{S} \circ \mathbb{C}(\mathfrak{g}_1 M) = \mathfrak{g}_1 \mathbb{S} \circ \mathbb{C} (M)$ for any $\mathfrak{g}_1 \in \mathfrak{G}_1$ and $M \in X$. Note that the $\mathbb{S}$ defined by (12) not only introduces the non-linearity in feature maps, but also converts the covariance $\mathfrak{g}_1'$ (from $\mathbb{C}$ w.r.t. rotation and flipping) into the more manageable equivariance $\mathfrak{g}_1$. In addition, the composition $\mathbb{S} \circ \mathbb{C}$ preserves the \emph{scaling covariance} of $\mathbb{C}$ due to the element-wise act of $\mathbb{S}$, i.e.,  $\mathbb{S} \circ \mathbb{C}(\mathfrak{g}_2 M) = \mathfrak{g}_2' \mathbb{S} \circ \mathbb{C} (M)$ for any $\mathfrak{g}_2 \in \mathfrak{G}_2$ and $M \in X$.

\begin{definition} \textbf{(Local pooling layer).}
	For the input feature map $M(i,j;k)$ with $\Omega  = \{ 1,2,...,{N_i}\}  \times \{ 1,2,...,{N_j}\} $ and ${\rm H} = {\mathfrak{C}^K}$, the local pooling layer $\mathbb{P}$ is defined as identity function:
	\begin{equation}
		\mathbb{P}M = M.
	\end{equation}
\end{definition}
\emph{Remark.} According to related researches, downsampling operations (e.g., local max pooling) of CNN will variously impair (translation) equivariance, i.e., an approximation $\mathbb{P}(\mathfrak{g}M) \simeq \mathfrak{g}\mathbb{P}(M)$, especially for larger pooling scales or deeper network architectures, implying a trade-off between computational complexity and representation equivariance. Since the proposed representation is one-shot without the large-scale training of typical CNN, we neglect downsampling operations and simply set  $\mathbb{P} = {\rm{id}}$ when the computational cost is acceptable. Alternatively, more elegant pooling designs with better tradeoffs between complexity and equivariance can be employed to define $\mathbb{P}$, as detailed in the paper by Zhang \cite{ref21}. It is straightforward that the composition $\mathbb{P} \circ \mathbb{S} \circ \mathbb{C}$ has the same representation properties of $\mathbb{S} \circ \mathbb{C}$ based on the Definition 3.

\begin{definition} \textbf{(Invariant layer).}
	For the input feature map $M(i,j;k)$ with $\Omega  = \{ 1,2,...,{N_i}\}  \times \{ 1,2,...,{N_j}\} $ and ${\rm H} = {\mathfrak{C}^K}$, the invariant layer $\mathbb{I}$ is defined channel-wise as \textbf{global invariant representations} with (1) and (2):
	\begin{equation}
		\mathbb{I}M = \mathcal{I} (\{  \left< M(i,j;k),{V_{nm}}({x_i},{y_j}) \right> \} ),
	\end{equation}
	where $\mathcal{I}$ is a special transform mapping image moments to global invariants, w.r.t. the symmetry group of interest $\mathfrak{G}_0 \subseteq  \mathfrak{G}_1 \times \mathfrak{G}_2$ and any $M \in X$.
\end{definition}
\emph{Remark.} In the Definition 4, we have not restricted $\mathcal{I}$ to a fixed formula, allowing the generality of the discussion in the following Section 3.3; its specific designs (w.r.t. considered applications of this paper) will be given in Section 4.1. Note that, with Definitions 1 $\sim$ 3 and Section 2.2, the $M$ form $\mathbb{P} \circ \mathbb{S} \circ \mathbb{C}$ with its cascade will basically preserve the geometric information of $f$, specifically the translation, rotation and flipping symmetries. Therefore, the idea of $\mathcal{I}$ (w.r.t. the deep feature map $M$) is very similar to the classical theory of moment invariants (w.r.t. the original image $f$), with a wide range of potential designs \cite{ref17}.

\begin{definition} \textbf{(Path).}
	From the Definitions 1 $\sim$ 4, we define a path of HIR as $p = ({\lambda _{[1]}},{\lambda _{[2]}},\cdots,{\lambda _{[L]}})$, where ${\lambda _{[z]}} = {(n,m,w)_{[z]}}$ specifies the parameters of the convolutional layer sorted by $z$. The HIR along a path $p$, ${{\cal R}_p}$, is defined as the following ordered cascading with corresponding parameters $p = ({\lambda _{[1]}},{\lambda _{[2]}},...,{\lambda _{[L]}})$:
	\begin{equation}
		{\cal R}_p  \triangleq \mathbb{I} \circ \mathbb{P}_{[L]} \circ \mathbb{S}_{[L]} \circ \mathbb{C}_{[L]} \circ \cdots \circ \mathbb{P}_{[1]} \circ \mathbb{S}_{[1]} \circ \mathbb{C}_{[1]}.
	\end{equation}
\end{definition}
\emph{Remark.} In the Definition 5, we further unify the global and local representation theories of moment invariants (Section 2) into a hierarchical representation framework. Note that the layers prior to the invariant layer $\mathbb{I}$ provide structure-preserving properties for the representation ${\cal R}_p$. Here, the global representation (15) is designed for image-level visual tasks, e.g., classification; as for pixel-level ones, e.g., segmentation, we can preserve the spatial dimensions by removing the last invariant layer. In the next section, the representation properties of Definition 5 will be analyzed explicitly.

\subsection{Representation Property}

In a typical CNN, the relationship between image information and learned representation is highly nonlinear and difficult to understand or predict. As for the HIR, we can explicitly give the following conclusions about the geometric symmetries between image and representation, implying good robustness and interpretability.

\begin{property} \textbf{(Equivariance for translation, rotation, and flipping).}
	For a representation unit $\mathbb{U} \triangleq \mathbb{P} \circ \mathbb{S} \circ \mathbb{C}$ with arbitrary parameters $\lambda$ (for the convolutional layer), any composition of $\mathbb{U}$ satisfy the joint equivariance for translation, rotation, and flipping (ignoring edge effects and resampling errors), i.e., the following identity holds:
	\begin{equation}
		  \mathbb{U}_{[L]} \circ \cdots \circ \mathbb{U}_{[2]} \circ \mathbb{U}_{[1]}(\mathfrak{g}_1 M)  \equiv \mathfrak{g}_1 \mathbb{U}_{[L]} \circ \cdots \circ \mathbb{U}_{[2]} \circ \mathbb{U}_{[1]}(M).
	\end{equation}
	for any composition length $L \ge 1$, any $\mathfrak{g}_1 \in \mathfrak{G}_1$ and $M \in X$, where $\mathfrak{G}_1$ is the translation/rotation/flipping symmetry group.
\end{property}

\begin{proof}
	First, let us examine the behavior of a representation unit $\mathbb{U}$ on $\mathfrak{G}_1$:
	\begin{equation}
		\begin{split}
		\mathbb{U}(\mathfrak{g}_1 M) & = \mathbb{P} \circ \mathbb{S} \circ \mathbb{C} (\mathfrak{g}_1 M)\\
		& = \mathbb{P} \circ \mathbb{S} \circ \mathfrak{g}_1' \mathbb{C} (M)\\
		& = \mathbb{P} \circ \mathfrak{g}_1 \mathbb{S} \circ \mathbb{C} (M)\\
		& = \mathfrak{g}_1 \mathbb{P} \circ \mathbb{S} \circ \mathbb{C} (M)\\
		& = \mathfrak{g}_1 \mathbb{U} (M),
		\end{split}
	\end{equation}
	where the first pass comes from the covariance of $\mathbb{C}$ for rotation and flipping, i.e., (6) and (7), and $\mathfrak{g}_1'$ is a predictable operation acting in the phase domain of $\mathbb{C}(M)$; the second pass comes from the specialization of $\mathbb{S}$ to the covariant $\mathfrak{g}_1'$ -- the magnitude operation removes the extra phase variations, leading to a pure equivariance $\mathfrak{g}_1$; the third pass comes from the identity function of $\mathbb{P}$, which becomes approximately equal when the downsampled $\mathbb{P}$ is used.
	
	Here, $\mathbb{U}(\mathfrak{g}_1 M) = \mathfrak{g}_1 \mathbb{U} (M)$ means that the representation unit $\mathbb{U}$ can be considered as an \emph{equivariant layer} for any $\mathfrak{g}_1 \in \mathfrak{G}_1$ and $M \in X$ -- in other words, the single $\mathbb{U}$ and $\mathfrak{g}_1$  operations on $M \in X$ are \emph{exchangeable}. Furthermore, with a notation $M_{[l]} \triangleq \mathbb{U}_{[l]} \circ \cdots \circ \mathbb{U}_{[1]}(M) = \mathbb{U}_{[l]} M_{[l-1]}$, we have $M_{[l]} \in X$ for any $l \in \{ 1,2,\cdots,L\} $. Therefore,  $\mathfrak{g}_1$ and any composition of $\mathbb{U}$ are exchangeable, implying the correctness of Property 1.
\end{proof}

\begin{property} \textbf{(Covariance for scaling).}
	For a representation unit $\mathbb{U}$, where the scale parameter of its convolutional layer is specified as $w$ with a notation $\mathbb{U}^w \triangleq \mathbb{P} \circ \mathbb{S} \circ \mathbb{C}^w$, any composition of $\mathbb{U}^w$ satisfy the covariance for scaling (ignoring edge effects and resampling errors), i.e., the following identity holds:
	\begin{equation}
		\begin{split}
		&\mathbb{U}^w_{[L]} \circ \cdots \circ \mathbb{U}^w_{[2]} \circ \mathbb{U}^w_{[1]}(\mathfrak{g}_2 M) \\
		&\equiv \mathfrak{g}_2' \mathbb{U}^w_{[L]} \circ \cdots \circ \mathbb{U}^w_{[2]} \circ \mathbb{U}^w_{[1]}(M)\\
		&= \mathfrak{g}_2 \mathbb{U}^{ws}_{[L]} \circ \cdots \circ \mathbb{U}^{ws}_{[2]} \circ \mathbb{U}^{ws}_{[1]}(M),
		\end{split}
	\end{equation}
	for any composition length $L \ge 1$, any $\mathfrak{g}_2 \in \mathfrak{G}_2$ and $M \in X$, where $\mathfrak{g}_2'$ is a predictable operation corresponding to $\mathfrak{g}_2$ with explicit form  $\mathfrak{g}_2'\mathbb{U}^w \triangleq \mathfrak{g}_2\mathbb{U}^{ws}$, $s$ is the scaling factor w.r.t. $\mathfrak{g}_2$, and $\mathfrak{G}_2$  is the scaling symmetry group.
\end{property}

\begin{proof}
	First, let us examine the behavior of a representation unit $\mathbb{U}^w$ on $\mathfrak{G}_2$:
	\begin{equation}
		\begin{split}
			\mathbb{U}^w(\mathfrak{g}_2 M) & = \mathbb{P} \circ \mathbb{S} \circ \mathbb{C}^w (\mathfrak{g}_2 M)\\
			& = \mathbb{P} \circ \mathbb{S} \circ \mathfrak{g}_2' \mathbb{C}^w (M)\\
			& = \mathbb{P} \circ \mathbb{S} \circ \mathfrak{g}_2 \mathbb{C}^{ws} (M)\\
			& = \mathbb{P} \circ \mathfrak{g}_2 \mathbb{S} \circ \mathbb{C}^{ws} (M)\\
			& = \mathfrak{g}_2 \mathbb{P} \circ \mathbb{S} \circ \mathbb{C}^{ws} (M)\\
			& = \mathfrak{g}_2 \mathbb{U}^{ws} (M)\\
			& = \mathfrak{g}_2' \mathbb{U}^w (M),
		\end{split}
	\end{equation}
	where the first pass comes from the covariance of $\mathbb{C}$ for scaling, i.e., (8), and $\mathfrak{g}_2'$ is a predictable operation acting in both the $\Omega$ domain (i.e., the same scaling $\mathfrak{g}_2$) and the $w$ domain (i.e., the factor $s$) of $\mathbb{C}^w (M)$; the second and third passes come from the element-wise act of $\mathbb{S}$ and the identity function of $\mathbb{P}$, respectively.

	Here, $\mathbb{U}^w(\mathfrak{g}_2 M) = \mathfrak{g}_2' \mathbb{U}^w (M)$ means that the representation unit $\mathbb{U}^w$ can be considered as an \emph{covariant layer} for any $\mathfrak{g}_2 \in \mathfrak{G}_2$ and $M \in X$ -- in other words, the single $\mathbb{U}^w$ and $\mathfrak{g}_2$ operations on $M \in X$ are \emph{exchaneable but with the parameter changing} of $ws$. Furthermore, we have ${M_{[l]}} \in X$ for any $l \in \{ 1,2,\cdots,L\} $. Therefore, $\mathfrak{g}_2$ and any composition of $\mathbb{U}^w$ are exchangeable while changing the scale parameter to $ws$, implying the correctness of Property 2.
\end{proof}

\begin{property} \textbf{(Hierarchical invariance).}
	For any composition of representation unit $\mathbb{U}$, it is practical to design a global invariant map $\mathcal{I}$ w.r.t. the symmetry group of interest $\mathfrak{G}_0 \subseteq  \mathfrak{G}_1 \times \mathfrak{G}_2$, due to the predictable geometric symmetries between the input image and deep feature map (at each intermediate layer) guaranteed by Properties 1 and 2. More specifically, with the Definition 4, we assume that there exists a $\mathcal{I}$ such that  $\mathbb{I}(\mathfrak{g}_0' M) = \mathbb{I}(M)$ for any $\mathfrak{g}_0 \in \mathfrak{G}_0$ and $M \in X$, i.e., invariance holds on one layer, where $\mathfrak{g}'$ is a predictable operation corresponding to $\mathfrak{g}$ and $\mathbb{U}$. Then we have following invariance:
	\begin{equation}
		\begin{split}
			 \mathbb{I}(\mathfrak{g}_0' M)_{[L]} \equiv \mathbb{I}M_{[L]},
		\end{split}
	\end{equation}
	holds for any composition length $L \ge 1$.
\end{property}

\begin{proof}
	We can rewrite (20) as:
	\begin{equation}
		\begin{split}
			\mathbb{I}(\mathfrak{g}_0 M)_{[L]} & = \mathbb{I}\circ \mathbb{U}_{[L]} \circ \cdots \circ \mathbb{U}_{[2]} \circ \mathbb{U}_{[1]}(\mathfrak{g}_0 M)\\
			& = \mathbb{I}(\mathfrak{g}_0' \mathbb{U}_{[L]} \circ \cdots \circ \mathbb{U}_{[2]} \circ \mathbb{U}_{[1]}(M))\\
			& = \mathbb{I} \circ \mathbb{U}_{[L]} \circ \cdots \circ \mathbb{U}_{[2]} \circ \mathbb{U}_{[1]}(M)\\
			& = \mathbb{I}M_{[L]},
		\end{split}
	\end{equation}
	where the first pass comes from Properties 1 and 2, note that $\mathfrak{g}_0 \in \mathfrak{G}_0 \subseteq  \mathfrak{G}_1 \times \mathfrak{G}_2$, $\mathfrak{g}_0'$ is related to $\mathfrak{g}_1$ and $\mathfrak{g}_2'$; the second pass comes from our assumption $\mathbb{I}(\mathfrak{g}_0' M) = \mathbb{I}(M)$ for any $\mathfrak{g}_0 \in \mathfrak{G}_0$ and $M \in X$, with ${M_{[l]}} \in X$ for any $l \in \{ 1,2,\cdots,L\}$.
\end{proof}

\subsection{Fast and Accurate Implementation}

Above, the core components of our representations, i.e., definitions and properties, have been formalized. In this section, we will complement the numerical implementation of HIR, especially the fast and accurate computations of Definition 1 from our previous work \cite{ref46}. Note that the discussion here is very general, with no restrictions on the specific definitions of the basis functions.

\begin{definition} \textbf{(Fast implementation).}
	Let us introduce the \textbf{convolution theorem} as a fast implementation of Definition 1, such that the spatial domain convolution of (9) can be converted to the following frequency domain product form \cite{ref46}:
	\begin{equation}
		\mathbb{C}M= {\mathcal{F}^{ - 1}}(\mathcal{F}(M(i,j;k)) \odot \mathcal{F}({(H_{nm}^w(i,j))^T})),
	\end{equation}
	where $\mathcal{F}$ is the Fourier transform and $\odot$ is the point-wise multiplication.
\end{definition}

\begin{property} \textbf{(Complexity analysis).}
	In the Definition 1, the (9) dominates the computational complexity due to the dense convolution. For the input feature map $M(i,j;k)$ with $\Omega  = \{ 1,2,\cdots,{N_i}\}  \times \{ 1,2,\cdots,{N_j}\} $ and ${\rm H} = {\mathfrak{C}^K}$, we assume that a set of $\mathbb{C}M$ needs to be computed, where scale parameter $w \in {S_w}$ with a fixed order $(n,m)$ and a fixed channel $k$, and denote the number of feature map samples as ${N_{ij}} = {N_i}{N_j}$ and the number of scale samples as ${N_w} = |{S_w}|$. With the Definition 6 and the Fast Fourier Transform (FFT), we can compute the set of $\mathbb{C}M$ in ${\mathcal{O}}({N_w}{N_{ij}}\log {N_{ij}})$ multiplications, as opposed to the complexity of ${\mathcal{O}}({N_w}{N_{ij}}{w_{\max }}^2)$ by the direct Definition 1, where ${w_{\max }}$ is the maximum scale in ${S_w}$. Note that the big difference between square and logarithmic growths in the complexity (removing the same terms), where the Definition 6 will exhibit batter efficiency when ${w_{\max }}$ is sufficiently large such that ${w_{\max }}^2 > \log {N_{ij}}$.
\end{property}

\begin{definition} \textbf{(Accurate implementation).}
	Let us introduce the \textbf{higher-order numerical integration} as an accurate implementation of Definition 1, such that the two-dimensional continuous integral of (11) can be converted to the following summation form \cite{ref46}:
	\begin{equation}
		h_{nm}^{uvw}(i,j) \simeq \sum\limits_{(a,b) \in {S_{ab}}} {{c_{ab}}{{(V_{nm}^{uvw}({x_a},{y_b}))}^*}\frac{{\Delta i\Delta j}}{{{w^2}}}},
	\end{equation}
	where the set of numerical integration samples ${S_{ab}}$ encodes the points $({x_a},{y_b}) \in {D_{ij}}$ and the corresponding weights ${c_{ab}}$, which are specified by a certain numerical integration strategy, such as Gaussian quadrature.
\end{definition}

\begin{property} \textbf{(Accurate analysis).}
	In the Definition 1, the (11) dominates the computational accuracy due to the continuous integration of complicated functions.  We assume that the $h_{nm}^{uvw}$ with a fixed order $(n,m)$ and position $(u,v)$ needs to be computed, and denote the number of numerical integration samples as ${N_{ab}} = |{S_{ab}}|$. The implementation based on the Definition 7 exhibits an approximation error of ${\mathcal{O}}({(\frac{{\Delta i\Delta j}}{{{w^2}}})^{{N_{ab}} + 1}})$. Note that when there is more than one sample within each pixel region, i.e., ${N_{ab}} > 1$, the Definition 7 will exhibit batter accuracy than the error of ${\mathcal{O}}({(\frac{{\Delta i\Delta j}}{{{w^2}}})^2})$ by the direct Definition 1 (zero-order approximation).
\end{property}

\subsection{Comparison with Related Work}

As a further explanation of Table 1, it is necessary to conclude this section and highlight the theoretical relationships with typical related works:

\begin{itemize}
	\item \emph{Traditional invariance}. Our work generalizes this theory by unifying the global and local invariant representations into a new framework of HIR. More specifically, we formalize layers $\mathbb{C}$, $\mathbb{S}$, and $\mathbb{P}$ based on the theory of local invariants \cite{ref46} (Definitions 1 $\sim$ 3), arguing the equivariance/covariance can be preserved across layers under a certain cascade (Properties 1 $\sim$ 2). We also formalize layer $\mathbb{I}$ based on the theory of global invariants \cite{ref17} (Definition 4), arguing the successes of global invariance for image domains can be directly generalized to equivariant/covariant deep feature domains (Property 3). Under our hierarchical invariance, classical global \cite{ref17} and local \cite{ref46} invariants can be considered as special cases, i.e., $\mathbb{I}f$ and $\mathbb{I} \circ \mathbb{S} \circ \mathbb{C} f$ (Definition 5).
	\item \emph{Traditional CNN}. Our work has a similar hierarchical architecture but with better properties in geometric symmetry, allowing for robust and interpretable image representations. More specifically, we introduce the discriminative design of CNN in our invariants, i.e., over-complete representation with deep cascading \cite{ref21}. On the other hand, we criticize typical CNN modules (Formulation 1), allowing fully transparent geometric symmetries across layers of our representation (Properties 1 $\sim$ 3). As a result, the proposed representation serves as an effective alternative to the highly black-box CNN in trustworthy tasks.
	\item \emph{Scattering networks}. Our work is more compact in achieving rotation invariance. As a main competitor, scattering networks are also based on deep cascading of explicit transforms (wavelets) \cite{ref24}, with similar concepts to our work. However, constructing rotation invariants from scattering networks is complicated, which requires parallel convolution and cross-channel pooling of multiple oriented wavelets; increasing the orientation sampling will result in an exponential growth of the complexity. Whereas our approach benefits from classical invariant theory, rotation invariance is continuous and one-shot (Property 1), providing better efficiency while easily enlarging the network size to improve the representation capacity.
	\item \emph{Equivariant networks}. Our work is non-learning while being more compact in achieving continuous and joint invariance. As a secondary competitor, equivariant networks are also guaranteed by group theory \cite{ref32}, with similar concepts to our work. However, the convolutional layers in equivariant networks are learned, leading to varying degrees of data dependence. In particular, it has a similar parallel structure to scattering networks, leading to exponential complexity and optimization challenges. Although equivariant networks are a very generic design, our approach provides better efficiency for continuous and joint invariance (Properties 1 $\sim$ 3), while easily enlarging the network size to improve the representation capacity.
\end{itemize}

\section{Hierarchical Invariance: Practice}

This section focuses on the practical aspects of the proposed HIR.

We first specify the above theory into a class of networks, involving more practical details about the topologies, layers, and parameters. We also discuss the domain adaptation strategies for practical scenarios, with feature/architecture selection or cascading learning module. Note that such efforts only serve as a feasible practice of hierarchical invariance theory, towards the experiments and applications in Section 5.

\subsection{Specifying the Architecture}

Starting from the Properties 1 $\sim$ 3, we propose a practical architecture for the hierarchical invariance w.r.t. the symmetry group of interest $\mathfrak{G}_0 = \mathfrak{G}_1 \times \mathfrak{G}_2$, with a tree topology and working at multiple scales.

\emph{Single-scale Networks.} Let us first present the topology on a single scale, i.e., all involved convolutional layers have a common scale parameter $w$, which exhibits invariance for $\mathfrak{G}_1$ (Properties 1 and 2). As shown in Fig. 2, we organize the set of paths as a tree-like network: 1) blue nodes denote the representation units $\mathbb{U}$ with different parameters $(n,m)$; 2) black nodes denote the identity function; 3) lines denote cascading relationships between nodes, where all nodes sorted by $l$ along their paths are plotted at the same level $l$ (sorted from top to bottom). Note that the feature map of each node will be fed into $\mathbb{I}$ for forming the invariants under this path, where the network representation is just the set of invariants under all paths. Here, the order parameter $(n,m)$ of the previous unit (blue) is always smaller than that of the subsequent ones (under a specific norm), so that the path exhibits an increasing trend in the order. With this design, the main information can be passed through the early nodes, and hence the subsequent nodes capture rich features. Also, the identity function (black) is introduced as a skip-connection trick, allowing the information to be passed to deeper nodes. In this paper, all units from the same level $l$ are specified separately from the set $\{ (n,m):n + m = l,(n,m) \in {\mathbb{N}^2}\} $, i.e., their orders are equal under the ${\ell _1}$ norm.

\emph{Multi-scale Networks.} Next, let us consider a multi-scale version of the above network with scale separation prior, extending the invariance to $\mathfrak{G}_0$ (Property 3). As shown in Fig. 3, a series of single-scale networks are introduced: 1) they have the same tree topology and same order parameter at corresponding nodes; 2) but each network has a different scale parameter, sampled from the set $\{ w:w = {2^t},t \in \mathbb{Z} \}$, where the scaling covariance (w.r.t. $w$) is transformed into a linear translation pattern (w.r.t. $t$) between multi-scale networks. The above network can derive a series of multi-scale representations of the image (with invariance for $\mathfrak{G}_1$), which are directly suitable for visual tasks with multi-scale physical structure (e.g., object detection). Further, we can derive scale-invariant representations (with invariance for $\mathfrak{G}_0$) under the Property 3, by pooling feature maps from a series of corresponding nodes at multiple scales. Note that, in practice, we cannot sample the scale completely and densely, and thus the above scaling invariance is restricted.

\begin{figure}[t]
	\centering
	\includegraphics[scale=0.43]{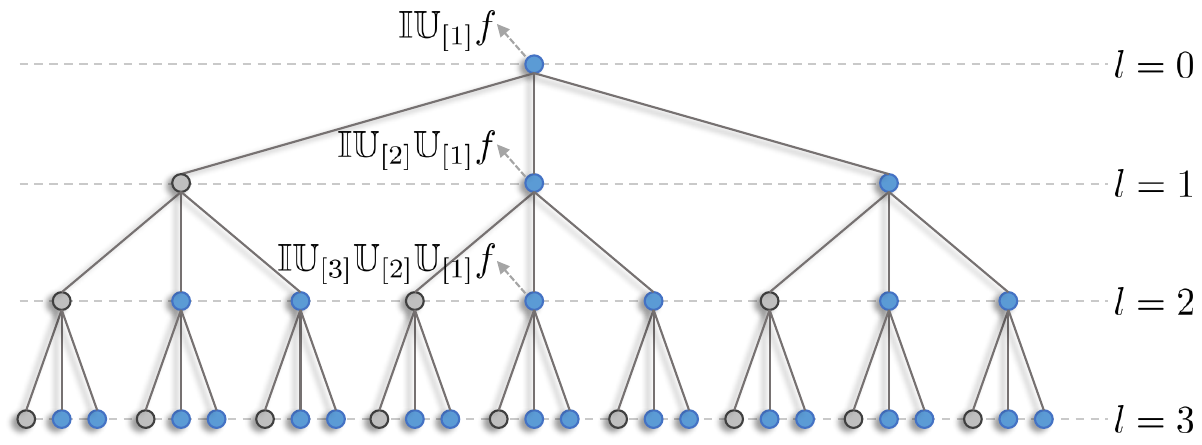}
	\centering
	\caption{A single-scale practice of HIR with the invariance for $\mathfrak{G}_1$. This tree-like HIR network encodes a set of paths, where blue and black nodes denote representation units (with different parameters) and identity function, respectively; lines denote cascading relationships between nodes.}
\end{figure}

\begin{figure}[t]
	\centering
	\includegraphics[scale=0.5]{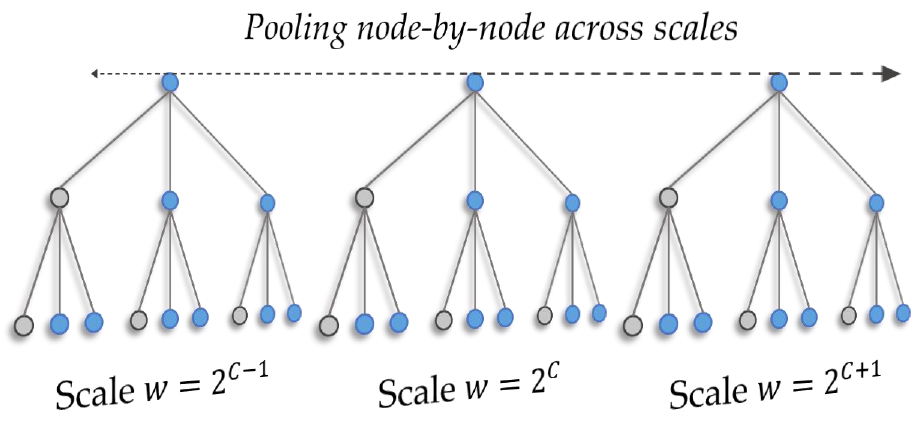}
	\centering
	\caption{A multi-scale practice of HIR with the invariance for $\mathfrak{G}_0$. This multi-scale HIR network is based on scale separation prior, where the scaling covariance is transformed into a linear translation pattern between multi-scale networks. One can derive scale-invariant representations by pooling feature maps from a series of corresponding nodes at multiple scales}
\end{figure}

\emph{Radial Basis Functions.} In our previous work \cite{ref49}, two generic classes of radial basis functions have been introduced, based on a family of harmonic functions:
\begin{equation}
	{R_n}(\alpha ,r) = \sqrt {\frac{{\alpha {r^{\alpha  - 2}}}}{{2\pi }}} \exp (\bm{j}2n\pi {r^\alpha }),
\end{equation}
and a family of polynomial functions:
\begin{equation}
	\begin{split}
		R_n^\alpha (p,q,r) = \sqrt {\frac{{\alpha {r^{\alpha q - 2}}{{(1 - {r^\alpha })}^{p - q}}(p + 2n) \Gamma (q + n) n!}}{{2\pi \Gamma (p + n) \Gamma (p - q + n + 1)}}} \\
		\times \sum\limits_{k = 0}^n {\frac{{{{( - 1)}^k}\Gamma (p + n + k){r^{\alpha k}}}}{{k!(n - k)!\Gamma (q + k)}}},
	\end{split}
\end{equation}
respectively, where the fractional parameter $\alpha \in {\mathbb{R}^+}$, the polynomial parameters $p,q \in {\mathbb{R}}$ must fulﬁll $p - q >  - 1$ and $q > 0$. Both classes of functions can be used to define   in the (2), satisfying the orthogonality condition in Section 2.1.

For the sake of simplicity, a family of cosine functions are chosen in all experiments and applications, as a special case of the (24):
\begin{equation}
	{R_n}(r) = \left\{ {\begin{array}{*{20}{l}}
			{\frac{1}{{\sqrt \pi  }}}&{n = 0}\\
			{\sqrt {\frac{2}{\pi }} \cos (n\pi {r^2})}&{n > 0}
	\end{array}} \right.,
\end{equation}
i.e., forming a hierarchical invariant version of the Polar Cosine Transform (PCT) \cite{ref50}. Note that we try to show the superiority of the hierarchical invariant framework itself, even if relying on naive (26).

\begin{figure*}[t]
	\centering
	\subfigure[Texture image dataset]
	{	\includegraphics[width=0.08\linewidth,height=0.08\linewidth]{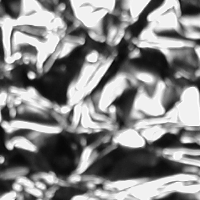}
		\includegraphics[width=0.08\linewidth,height=0.08\linewidth]{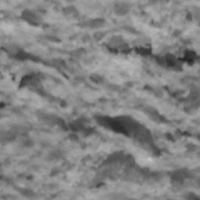}
		\includegraphics[width=0.08\linewidth,height=0.08\linewidth]{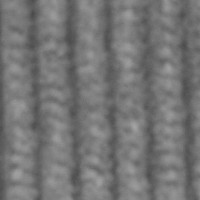}
		\includegraphics[width=0.08\linewidth,height=0.08\linewidth]{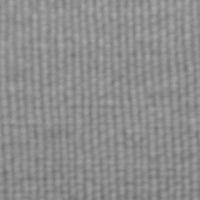}
		\includegraphics[width=0.08\linewidth,height=0.08\linewidth]{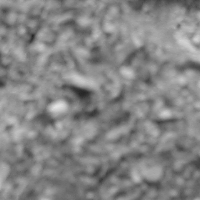}
		\includegraphics[width=0.08\linewidth,height=0.08\linewidth]{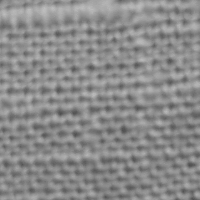}
		\includegraphics[width=0.08\linewidth,height=0.08\linewidth]{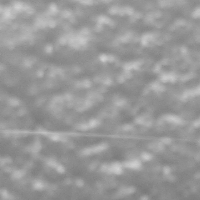}
		\includegraphics[width=0.08\linewidth,height=0.08\linewidth]{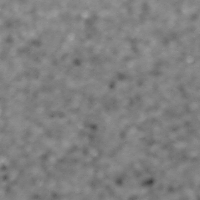}
		\includegraphics[width=0.08\linewidth,height=0.08\linewidth]{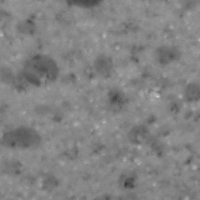}
		\includegraphics[width=0.08\linewidth,height=0.08\linewidth]{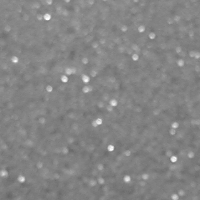}
	}
	
	\subfigure[Digit image dataset]
	{	\includegraphics[width=0.08\linewidth,height=0.08\linewidth]{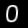}
		\includegraphics[width=0.08\linewidth,height=0.08\linewidth]{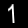}
		\includegraphics[width=0.08\linewidth,height=0.08\linewidth]{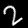}
		\includegraphics[width=0.08\linewidth,height=0.08\linewidth]{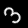}
		\includegraphics[width=0.08\linewidth,height=0.08\linewidth]{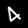}
		\includegraphics[width=0.08\linewidth,height=0.08\linewidth]{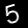}
		\includegraphics[width=0.08\linewidth,height=0.08\linewidth]{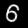}
		\includegraphics[width=0.08\linewidth,height=0.08\linewidth]{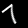}
		\includegraphics[width=0.08\linewidth,height=0.08\linewidth]{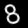}
		\includegraphics[width=0.08\linewidth,height=0.08\linewidth]{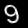}
	}
	
	\subfigure[Parasite image dataset]
	{	\includegraphics[width=0.08\linewidth,height=0.08\linewidth]{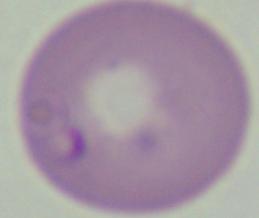}
		\includegraphics[width=0.08\linewidth,height=0.08\linewidth]{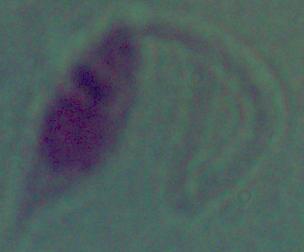}
		\includegraphics[width=0.08\linewidth,height=0.08\linewidth]{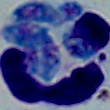}
		\includegraphics[width=0.08\linewidth,height=0.08\linewidth]{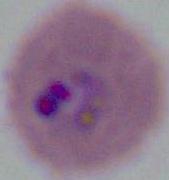}
		\includegraphics[width=0.08\linewidth,height=0.08\linewidth]{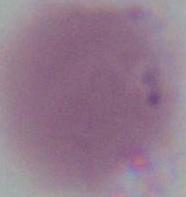}
		\includegraphics[width=0.08\linewidth,height=0.08\linewidth]{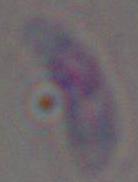}
		\includegraphics[width=0.08\linewidth,height=0.08\linewidth]{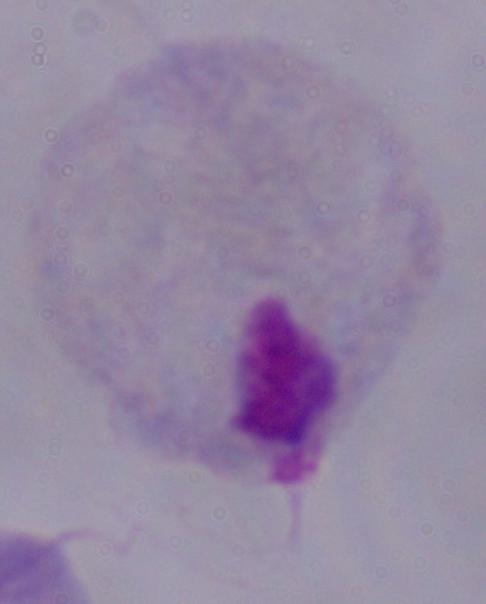}
		\includegraphics[width=0.08\linewidth,height=0.08\linewidth]{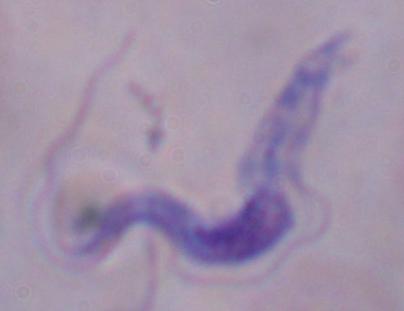}
	}
	
	\centering
	\caption{Illustration for the datasets from the computer vision and pattern recognition experiments.}
\end{figure*}

\emph{Invariant Layer.} In the monograph \cite{ref51} and our previous work \cite{ref17}, a number of strategies for directly constructing global invariants in image domains have been presented. They can be naturally used to define $\mathcal{I}$ in (14), with the equivariant or covariant behavior of deep feature maps (Properties 1 $\sim$ 3). In all experiments and applications of this paper, a class of global invariants is concisely designed based on frequency pooling.

Regarding (14), we first let the Fourier basis be ${V_{nm}}({x_i},{y_j})$. Note that the Fourier Transform (FT) is highly understood in the signal processing community and can be considered a good foundation for interpretability. Then, based on the order/frequency sampling of the FT $(n,m) \in {[ - K,K]^2}$, we define $\mathcal{I}$ as a frequency-band integral in the polar system:
\begin{equation}
	\begin{split}
	&{\cal I}(\{  \left< M,{V_{nm}} \right> \} )\\
	&\triangleq \{ {I_i} = \sum\limits_{(n,m) \in {{\cal B}_i}} {\{ | \left< M,{V_{nm}} \right> |\} :i = 1,2,\cdots,{\# _B}} \},
	\end{split}
\end{equation}
where ${{\cal B}_i} = \{ (n,m):\sqrt 2 K(i - 1)/{\# _B}| \le ||(n,m)|{|_2} \le \sqrt 2 Ki/{\# _B}\} $ is the $i$-th frequency band under the ${\ell _2}$ norm, with the number of bands ${\# _B}$.

Here, we can state that the above feature vector $\{ {I_i}:i = 1,2,...,{\# _B}\} $ directly satisfies the invariance for $\mathfrak{G}_1$, in light of Property 1 and the translation, rotation and flipping properties of FT. As for scaling, ${\cal I}$ is compatible with both single-scale and multi-scale networks: 1) regarding the single-scale case, a certain degree of robustness is provided for $\mathfrak{G}_2$ (at least up to the bandwidth), in light of Property 2 and the scaling property of FT; 2) regarding the multi-scale case, the scaling covariance has been eliminated before feeding into ${\cal I}$, and thus will satisfy the joint invariance for $\mathfrak{G}_0 = \mathfrak{G}_1 \times \mathfrak{G}_2$.

Note that the well-known average pooling is in fact a special case of (27), with $K = 0$ and ${\# _B} = 1$. Our frequency-band integral ${\cal I}$ can be regarded as a generic design of global pooling, with comprehensive consideration on interpretability, invariance, and discriminability.

\subsection{Empowering the Data Adaptability}

Due to the hand-crafted nature of HIR, a fixed set of their features is not adaptive to the data distribution. For larger-scale vision tasks, we propose following data adaptability strategies, allowing our invariants to reach a similar discriminability level of learning representations. Note that unlike typical (under)-complete invariants, our invariants exhibit a high level of over-completeness due to the local and hierarchical structure, which is the foundation for data adaptability strategies.

\emph{Feature/Architecture Selection.} Discriminative features for a given task can be formed in a selection-based manner, inspired by Neural Architecture Search (NAS) \cite{ref52}. First, we can construct a large-scale tree-like network (going deeper or wider), covering a wide set of paths and parameters, analogous to the notion of supernet in NAS \cite{ref53}. Then, with the training set under a given task, we can perform correlation analysis of features and labels for ranking discriminative features (as well as the corresponding paths), analogous to the phase of architecture sampling and evaluation in NAS \cite{ref53}. With the above analysis, we can greatly simplify the initial supernet such that the paths cover top-ranked features for applications, allowing our representations to be task-discriminative.

\emph{Cascading Learning Module.} Discriminative features for a given task can also be formed in a learning-based manner, inspired by Hybrid Representation Learning (HRL) \cite{ref27}. The main idea is to replace shallow layers of learning CNN with fixed HIR, such that discriminative features are formed in a space with geometric symmetries. According to related justifications \cite{ref27}, this strategy is able to achieve a discriminability level rivaling typical CNN on large-scale classification benchmarks, while exhibiting significantly better training compactness. In all experiments and applications of this paper, we still employ the feature/architecture selection strategy to show the superiority of the hierarchical invariant framework itself, taking also into account that the cascading CNN weakens invariance and interpretability to some extent.

\begin{table}[t]
	\centering
	\caption{Classification Scores (\%) and Runtime (Second) for Different Representations on a Small-scale Texture Benchmark.}
	\resizebox{\linewidth}{!}{
	\begin{tabular}{ccccccccc}
		\toprule
		\multirow{2}{*}{Method} & \multirow{2}{*}{\tabincell{c}{Time\\GPU\dag}} & \multicolumn{3}{c}{\textbf{Original}} & \multicolumn{3}{c}{\textbf{Orien. \& Flip.}} \\
		&         & Pre. & Rec. & F1 & Pre. & Rec. & F1 \\
		\midrule
		\emph{Classical:} &             &       &       &       &       &       &  \\
		\midrule
		Cosine & 5           & 70.74 & 67.50  & 66.85 & 69.65 & 66.25 & 65.30 \\
		Wavelet & 6            & 69.43 & 64.38 & 64.68 & 62.34 & 58.13 & 57.82 \\
		Kraw. & 5            & 70.67 & 67.50  & 66.30  & 64.41 & 60.00    & 59.55 \\
		\midrule
		\emph{Learning:} &              &       &       &       &       &       &  \\
		\midrule
		SimpleNet & 52\dag & 70.33 & 67.50  & 67.09 & 54.63 & 43.13 & 41.31 \\
		SimpleNet+ & 52\dag & 46.93 & 49.38 & 46.06 & 47.18 & 48.13 & 44.93 \\
		AlexNet & 42\dag & 98.82 & 98.75 & 98.75 & 91.69 & 91.25 & 91.28 \\
		AlexNet+ & 41\dag & 87.61 & 84.38 & 84.05 & 88.37 & 85.63 & 85.76 \\
		VGGNet & 266\dag & 99.41 & 99.38 & 99.37 & 92.18 & 91.25 & 91.37 \\
		VGGNet+ & 609\dag & 91.34 & 90.00  & 89.81 & 92.15 & 91.25 & 91.08 \\
		\midrule
		\emph{Invariant:} &             &       &       &       &       &       &  \\
		\midrule
		ScatterNet & 42           & 98.89 & 98.75 & 98.75 & 84.98 & 83.13 & 83.08 \\
		HIR   & 27          & 96.98 & 96.88 & 96.87 & 96.32 & 96.25 & 96.23 \\
		\bottomrule
	\end{tabular}
}
\end{table}%

\section{Experiments and Applications}

In this section, we will comprehensively evaluate the discriminability, robustness, and efficiency of HIR, covering simulation experiments in Section 5.1 and real-world applications in Section 5.2. Here, the main aim is for examining the representation properties promised in previous sections, as well as positioning its discriminative power in the era of deep learning.

With Sections 4 and 5, we implement a code repository for HIR in \texttt{https://github.com/ShurenQi/HIR}. All experiments/applications are executed in Matlab R2023a under Microsoft Windows environment, based on 2.90-GHz CPU, RTX-3060 GPU, and 16-GB RAM.

\subsection{Computer Vision and Pattern Recognition}

We perform classification experiments with HIR on typical sets of texture, digit, and parasite images, benchmarking its representation capabilities. Note that this series of simulation experiments examines the properties promised by our theory under diverse task scales and geometric variants, also with comparisons to a range of hand-crafted and learning representations.

With the practice of Sections 4, our HIR is implemented here as a single-scale network scale parameter $w = 10$ and composition length $L = 6$; its invariant layer (27) is specialized to the average pooling, with $K = 0$ and ${\# _B} = 1$, for a fair comparison with the deep representations by average pooling. Note that the adaptability strategies of Section 4.2 are not employed here, for a direct assessment of its discriminative power. All features are fed into a PCA classifier, trained on features of the training set. Unless otherwise stated, the training and testing sets are formed without any crossover by random sampling at 80\% and 20\% ratios on the original dataset, respectively.

The competing representations involved here can be summarized as follows:
\begin{itemize}
	\item Classical complete representation: discrete cosine transform as a global representation;
	\item Classical over-complete representation: discrete wavelet transform \cite{ref54} and Krawtchouk moments \cite{ref55} as local representations, with different time-frequency resolutions;
	\item Advanced over-complete representations: 1) typical CNNs, i.e., direct-learning CNN (denoted as SimpleNet), transfer-learning AlexNet \cite{ref56} and VGGNet \cite{ref57}, with also data augmentation (denoted as ‘+’); 2) invariant CNNs, i.e., scattering networks \cite{ref24} and our HIR. Broadly speaking, they can all be considered as a class of hierarchical invariant representations, but with different levels of invariance, where typical CNNs are invariant only to translations.
\end{itemize}

\begin{table}[t]
	\centering
	\caption{Classification Scores (\%) and Runtime (Second) for Different Representations on a Medium-scale Digit Benchmark.}
	\resizebox{\linewidth}{!}{
		\begin{tabular}{ccccccccc}
			\toprule
			\multirow{2}{*}{Method} & \multirow{2}{*}{\tabincell{c}{Time\\GPU\dag}} & \multicolumn{3}{c}{\textbf{Original}} & \multicolumn{3}{c}{\textbf{Trans. \& Rota.}} \\
			&       & Pre. & Rec. & F1 & Pre. & Rec. & F1 \\
			\midrule
			\emph{Classical:} &              &       &       &       &       &       &  \\
			\midrule
			Cosine & 15           & 45.68  & 45.35  & 45.43  & 32.50  & 31.20  & 30.83  \\
			Wavelet & 16           & 67.11  & 66.75  & 66.75  & 38.31  & 35.25  & 35.44  \\
			Kraw. & 15           & 71.73  & 69.85  & 69.69  & 27.05  & 26.30  & 25.83  \\
			\midrule
			\emph{Learning:} &              &       &       &       &       &       &  \\
			\midrule
			SimpleNet & 535\dag     & 98.60  & 98.60  & 98.60  & 35.42  & 33.50  & 33.72  \\
			SimpleNet+ & 551\dag     & 52.70  & 48.90  & 48.82  & 54.26  & 50.75  & 50.78  \\
			AlexNet & 393\dag     & 100  & 100  & 100  & 66.18  & 64.45  & 64.27  \\
			AlexNet+ & 392\dag     & 93.07  & 92.00  & 91.80  & 94.23  & 93.10  & 92.93  \\
			VGGNet & 3610\dag     & 100  & 100  & 100  & 70.74  & 70.25  & 69.93  \\
			VGGNet+ & 7731\dag     & 95.98  & 95.70  & 95.68  & 95.53  & 95.20  & 95.13  \\
			\midrule
			\emph{Invariant:} &              &       &       &       &       &       &  \\
			\midrule
			ScatterNet & 115          & 98.96  & 98.95  & 98.95  & 57.20  & 56.95  & 56.23  \\
			HIR   & 57           & 97.48  & 97.45  & 97.45  & 95.05  & 94.95  & 94.98  \\
			\bottomrule
		\end{tabular}
	}
\end{table}%

\subsubsection{Texture}

As shown in Fig. 4, the experiment is executed on dataset KTH-TIPS\footnote{https://www.csc.kth.se/cvap/databases/kth-tips/index.html}, a typical benchmark for texture image classification. This dataset has 10 classes, each containing 81 instances, the total size is $10 \times 81 = 810$, and hence is considered as a small-scale vision problem.

As shown in Table 2, we list performance scores of the competing representations on this benchmark, as well as the elapsed time, i.e., CPU featuring time or GPU training time. Besides this direct protocol on the original dataset, we also consider testing image variants with random orientation (w.r.t. $\{0, 90, 180, 270\}$ degree) or flipping (w.r.t. $x$ or $y$ axis).
\begin{itemize}
	\item The classical (over-)complete representations fail to achieve a satisfactory level of discriminability, even in the direct protocol of such small-scale benchmark.
	\item The learning CNN family achieves significantly higher scores due to its over-complete and data-adaptive properties, especially the AlexNet and VGGNet with large-scale pre-training and transfer learning. Whereas, the SimpleNet performs relatively poorly, indicating the sensitivity of learning to network size and training strategy. Under the variant protocol, they exhibit a significant performance degradation, suggesting the learned features lack invariance w.r.t. natural geometric variations of texture. After introducing the augmented training, the CNN scores become more stable, but at the cost of discriminability. A potential reason for this phenomenon is the small amount of training data. Moreover, the computational cost is considerable for this small-scale problem, and a certain training instability is observed.
	\item The scattering networks provide a high level of discriminability and robustness without feature training and data augmentation, indicating the success of extending classical wavelets to deep representations.
	\item Our work further extends such success: the HIR achieves a similar level of discriminability as the learning CNN family, while exhibiting superior robustness in the variant protocol than all competing representations. In particular, such representation success build on our compact and efficient framework, with lower runtimes than scattering networks and learning CNN family.
\end{itemize}

\subsubsection{Digit}

As shown in Fig. 4, the experiment is executed on a digit dataset\footnote{https://ww2.mathworks.cn/help/deeplearning/ug/data-sets-for-deep-learning.html} for classification similar to the MNIST benchmark. This dataset has 10 classes from ‘0’ to ‘9’, each containing 1000 instances with rich font differences and geometric distortions, the total size is $10 \times 1000 = 10000$, and hence is considered as a medium-scale vision problem.

As shown in Table 3, we list performance scores and elapsed times of the competing representations on this benchmark. Besides this direct protocol, we also consider testing image variants with random translation (w.r.t. $-2 \sim 2$ pixels in axial directions) and random rotation (w.r.t. $0 \sim 360$ degrees around the center). 

\begin{itemize}
	\item As the problem size increases, the complete Cosine exhibits a significant degradation, while the over-complete Wavelet and Krawtchouk are more stable, revealing the role of over-completeness in the discriminability. Regarding robustness, such representations all fail against translation and rotation variants, implying the challenging nature of this protocol.
	\item In general, the learning CNN family continues the performance in Section 5.1.1, further confirming the sensitivity of learning to network size, training strategy, and geometric variants. Its robustness is significantly increased after augmented training. However, theoretically, the resulting robustness is not guaranteed for unseen data distributions (even for similar variants with unseen parameters). One can note the rapid expansion of computational cost: the transfer learning of VGGNet even takes $\sim$ 2 GPU hours.
	\item The handcrafted scattering networks still provide a good level of discriminability here, further validating its success. However, it exhibits unsatisfactory scores for translation and rotation variants, even lower than the CNN without augmentation, failing to achieve the expected robustness.
	\item The HIR significantly outperforms the main competitor, i.e., scattering networks. Despite the increased problem size, it still achieves a similar level of discriminability as the learning CNN family, also under the constraints of invariance and compactness. Note that our HIR is the only method exhibiting confusion between classes ‘6’ and ‘9’ in the direct training, while achieving $\sim$ 100\% scores for the rest. This phenomenon is in line with the rotation invariance and discriminative power expected by our theory.
	
\end{itemize}

\begin{table}[t]
	\centering
	\caption{Classification Scores (\%) and Runtime (Second, for Train./Test. = 8/2) for Different Representations on a Large-scale Parasite Benchmark.}
	\resizebox{\linewidth}{!}{
	\begin{tabular}{ccccccccc}
		\toprule
		\multirow{2}{*}{Method} & \multirow{2}{*}{\tabincell{c}{Time\\GPU\dag}}  & \multicolumn{3}{c}{\textbf{Train./Test. = 8/2}} & \multicolumn{3}{c}{\textbf{Train./Test. = 1/9}} \\
		&        & Pre. & Rec. & F1 & Pre. & Rec. & F1 \\
		\midrule
		\emph{Classical:} &       &       &       &       &       &       &  \\
		\midrule
		Cosine & 37    & 36.19  & 32.60  & 29.85  & 49.40  & 41.97  & 43.80  \\
		Wavelet & 39    & 41.68  & 45.20  & 41.79  & 53.69  & 47.97  & 49.27  \\
		Kraw. & 42    &   66.56  & 69.49  & 67.21  & 71.60  & 57.88  & 61.10  \\
		\midrule
		\emph{Learning:} &        &       &       &       &       &       &  \\
		\midrule
		SimpleNet & 2244\dag   & 90.15  & 89.25  & 89.65  & 84.51  & 76.14  & 78.84  \\
		AlexNet & 1796\dag   & 98.87  & 98.40  & 98.63  & 95.92  & 94.69  & 95.27  \\
		VGGNet & 9184\dag   & 99.24  & 98.97  & 99.11  & 97.95  & 97.37  & 97.65  \\
		\midrule
		\emph{Invariant:} &      &       &       &       &       &       &  \\
		\midrule
		ScatterNet & 1277    & 68.41  & 69.71  & 67.55  & 72.52  & 63.30  & 65.70  \\
		HIR   & 823      & 88.73  & 92.18  & 90.10  & 91.26  & 88.76  & 89.85  \\
		\bottomrule
	\end{tabular}
}
\end{table}%

\subsubsection{Parasite}

As shown in Fig. 4, the experiment is executed on micrographic dataset\footnote{https://data.mendeley.com/datasets/38jtn4nzs6/3}, a typical benchmark for parasite image classification. This dataset has 6 parasite classes and 2 host classes, with real-world diversity regarding imaging, background, morphology, and geometry, the total size is 34298, and hence is considered as a large-scale vision problem.

As shown in Table 4, we list performance scores and elapsed times of the competing representations on this benchmark. Note that we also consider a protocol with different training-testing ratios to analyze the data dependence and sample efficiency.

\begin{itemize}
	\item In this large-scale problem, the scores of the classical representations drop further, implying a limited level of discriminability. On the other hand, their performance is relatively stable when training samples are reduced, and even better in the $1/9$ case, indicating a good efficiency.
	\item In the learning CNN family, the direct-learning SimpleNet exhibits a clear data dependence. Specifically, it achieves $\sim$90\% scores in the $8/2$ case (similar to HIR), while the scores drop significantly in the $1/9$ case (below than HIR). In contrast, the AlexNet and VGGNet achieve good discriminability and stability in the $1/9$ case, revealing that the transfer strategy effectively inherits the pre-training prior on ImageNet. On the other hand, the cost of pre-training and transfer learning is still considerable, without guaranteed robustness or adaptability for a given data domain.
	\item Despite outperforming the original wavelets, scattering networks fail to provide a competitive discriminability in the era of deep learning. Here, the common failure of such hand-crafted representations on larger-scale discriminability can be regarded as important evidence for our motivation.
	\item The HIR achieves a SimpleNet-level discriminability, outperforming our competitor scattering networks significantly. Also, the HIR is not sensitive to the reduction of training samples, outperforming the learning CNN family in data dependence and sample efficiency. Note that the discriminability of the fixed features from HIR is still lower than the transfer learning with large-scale pre-training. Therefore, in the next applications, the HIR features will be empowered with data adaptability strategies in Section 4.2.

\end{itemize}

\begin{figure*}[t]
	\centering
	\subfigure[Adversarial perturbation image dataset]
	{	\includegraphics[width=0.12\linewidth,height=0.12\linewidth]{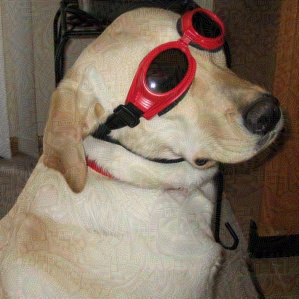}
		\includegraphics[width=0.12\linewidth,height=0.12\linewidth]{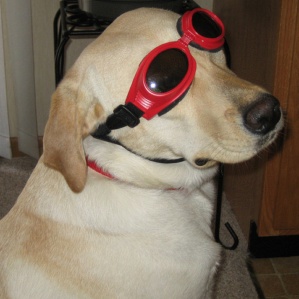}
		\includegraphics[width=0.12\linewidth,height=0.12\linewidth]{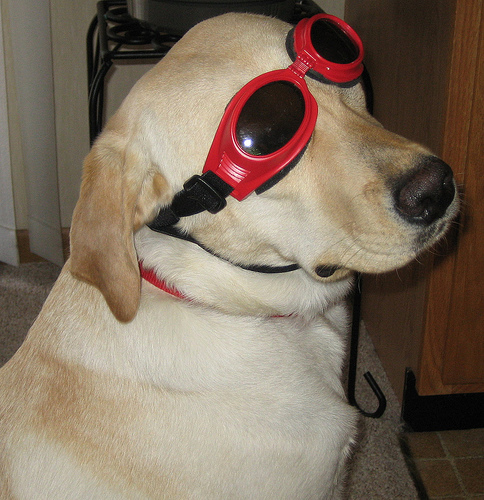}
		\includegraphics[width=0.12\linewidth,height=0.12\linewidth]{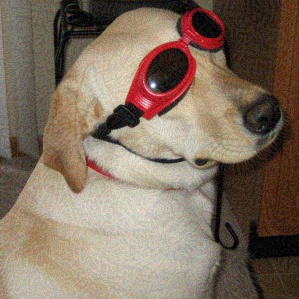}
		\includegraphics[width=0.12\linewidth,height=0.12\linewidth]{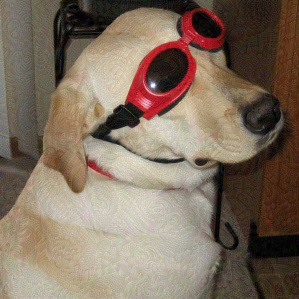}
		\includegraphics[width=0.12\linewidth,height=0.12\linewidth]{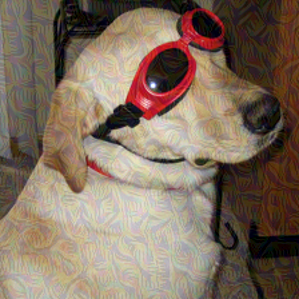}
	}
	
	\subfigure[AIGC image dataset]
	{	\includegraphics[width=0.12\linewidth,height=0.12\linewidth]{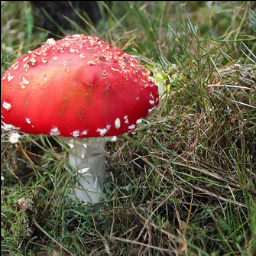}
		\includegraphics[width=0.12\linewidth,height=0.12\linewidth]{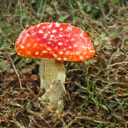}
		\includegraphics[width=0.12\linewidth,height=0.12\linewidth]{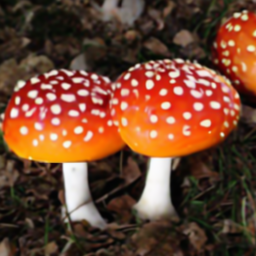}
		\includegraphics[width=0.12\linewidth,height=0.12\linewidth]{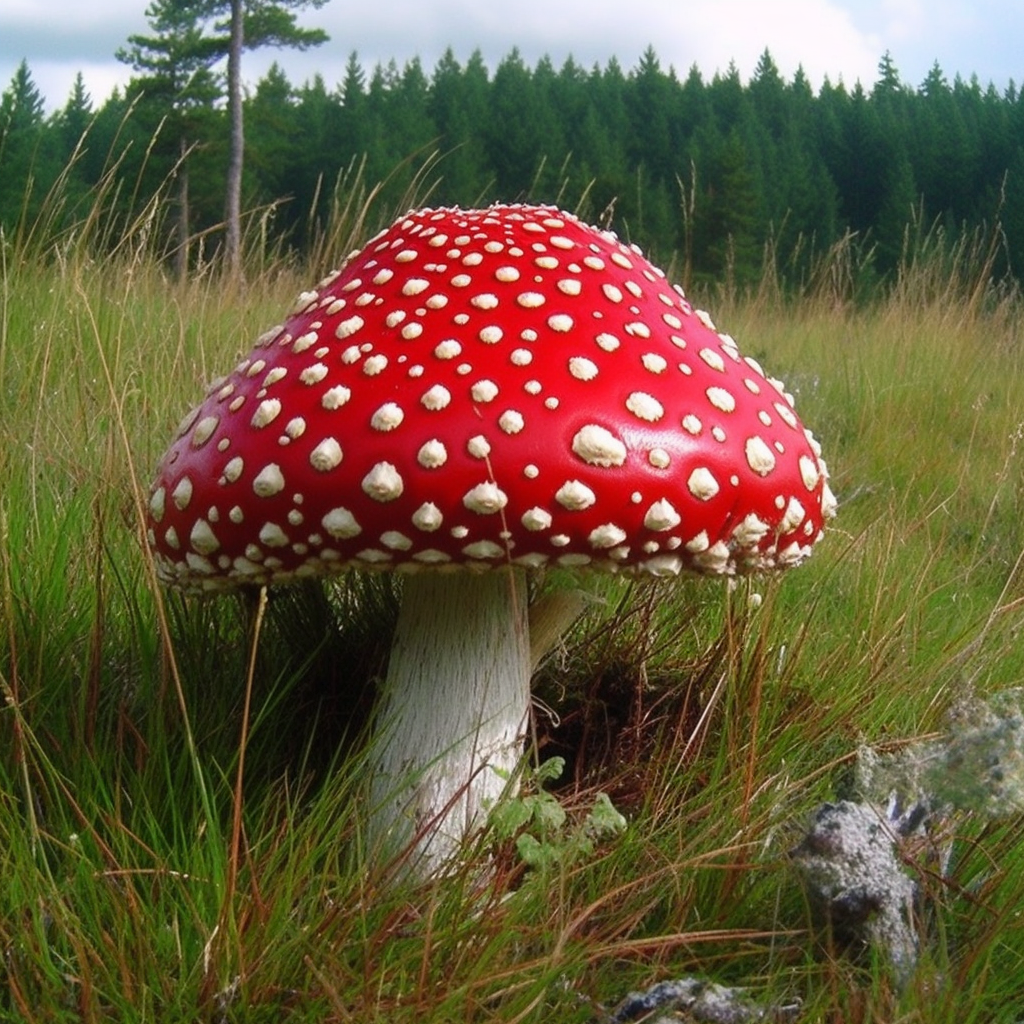}
		\includegraphics[width=0.12\linewidth,height=0.12\linewidth]{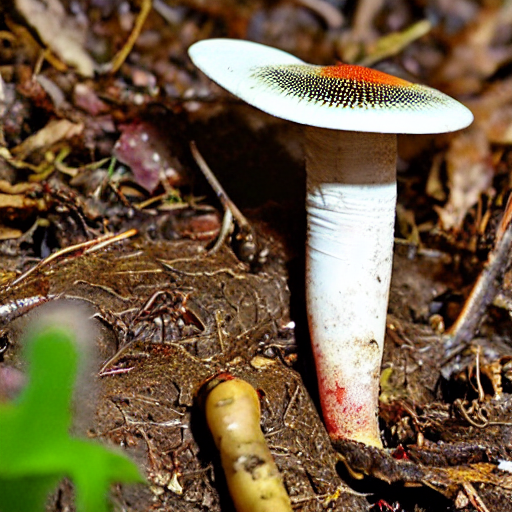}
		\includegraphics[width=0.12\linewidth,height=0.12\linewidth]{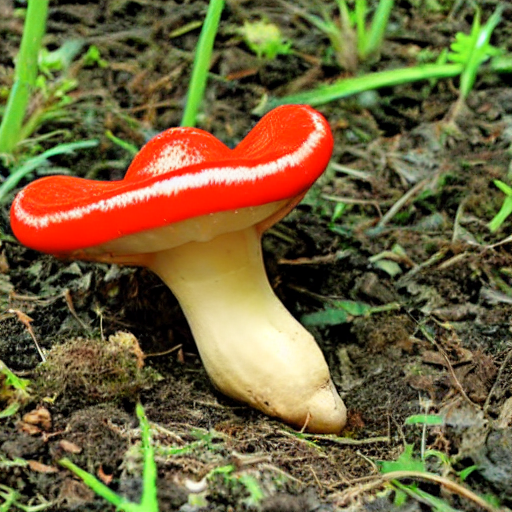}
		\includegraphics[width=0.12\linewidth,height=0.12\linewidth]{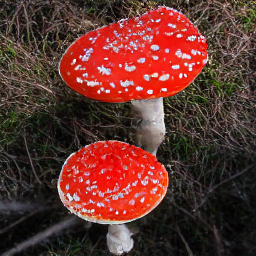}
		\includegraphics[width=0.12\linewidth,height=0.12\linewidth]{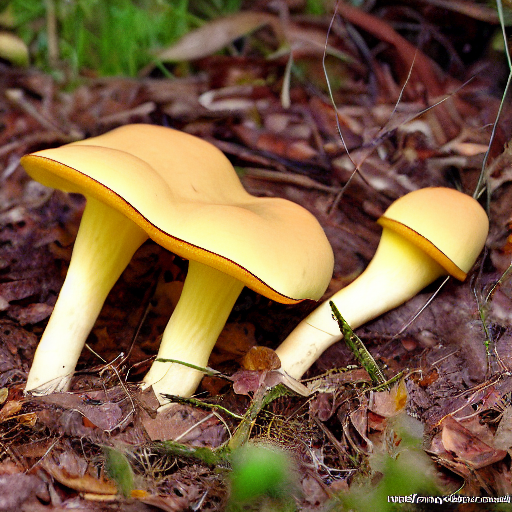}
	}
	
	\centering
	\caption{Illustration for the datasets from the digital forensic and forgery detection experiments.}
\end{figure*}

\begin{table*}[t]
	\centering
	\caption{Adversarial Perturbation Forensic Scores (F1, \%) for Different Representations w.r.t. Various Types of Perturbations.}
	\begin{tabular}{ccccccccc}
		\toprule
		Method & BIM & CW & DAmage & FGSM & PGD & UP & \textbf{Average} & \textbf{Worst} \\
		\midrule
		\emph{Classical:} &       &       &       &       &       &       &       &  \\
		\midrule
		Cosine NN & 34.63  & 33.19  & 90.78  & 39.80  & 34.69  & 2.22  & 39.22  & 2.22  \\
		Cosine SVM & 79.57  & 83.34  & 97.26  & 78.24  & 79.22  & 96.68  & 85.72  & 78.24  \\
		Wavelet NN & 0.00  & 0.00  & 0.00  & 0.00  & 0.00  & 0.00  & 0.00  & 0.00  \\
		Wavelet SVM & 72.83  & 82.09  & 97.77  & 78.21  & 71.80  & 95.87  & 83.10  & 71.80  \\
		Krawtchouk NN & 66.43  & 66.49  & 90.86  & 66.43  & 66.44  & 0.00  & 59.44  & 0.00  \\
		Krawtchouk SVM & 0.00  & 55.87  & 0.00  & 56.44  & 0.00  & 70.37  & 30.45  & 0.00  \\
		\midrule
		\emph{Learning:} &       &       &       &       &       &       &       &  \\
		\midrule
		SimpleNet & 4.24  & 3.24  & 92.13  & 49.89  & 33.13  & 99.86  & 47.08  & 3.24  \\
		AlexNet & 90.20  & 72.72  & 96.63  & 94.61  & 90.91  & 98.45  & 90.59  & 72.72  \\
		VGGNet & 96.04  & 62.50  & 99.08  & 98.12  & 96.99  & 99.15  & 91.98  & 62.50  \\
		GoogLeNet & 90.29  & 80.04  & 97.09  & 95.29  & 89.94  & 98.75  & 91.90  & 80.04  \\
		ResNet & 90.22  & 75.59  & 97.35  & 94.66  & 90.17  & 98.40  & 91.07  & 75.59  \\
		DenseNet & 98.93  & 90.19  & 99.34  & 99.23  & 98.85  & 99.76  & 97.72  & 90.19  \\
		InceptionNet & 98.70  & 85.14  & 97.38  & 97.32  & 98.66  & 99.41  & 96.10  & 85.14  \\
		MobileNet & 92.51  & 82.67  & 97.37  & 96.81  & 92.10  & 98.19  & 93.27  & 82.67  \\
		\midrule
		\emph{Invariant:} &       &       &       &       &       &       &       &  \\
		\midrule
		ScatterNet NN & 81.30  & 70.23  & 95.27  & 91.17  & 82.65  & 94.64  & 85.88  & 70.23  \\
		ScatterNet SVM & 84.40  & 69.49  & 96.77  & 90.57  & 83.86  & 95.12  & 86.70  & 69.49  \\
		HIR NN & 89.66  & 84.92  & 98.89  & 93.26  & 90.08  & 97.78  & 92.43  & 84.92  \\
		HIR SVM & 92.30  & 89.10  & 99.30  & 95.96  & 91.60  & 98.93  & 94.53  & 89.10  \\
		\bottomrule
	\end{tabular}%
	
\end{table*}%

\begin{table}[t]
	\centering
	\caption{Adversarial Perturbation Forensic Scores (\%) for Different Representations on a Real-world (Hybrid) Benchmark.}
	\begin{tabular}{ccccccc}
		\toprule
		\multirow{2}{*}{Method} & \multicolumn{3}{c}{\textbf{Train./Test. = 5/5}} & \multicolumn{3}{c}{\textbf{Train./Test. = 1/9}} \\
		&   Pre. & Rec. & F1 & Pre. & Rec. & F1 \\
		\midrule
		\emph{Classical:} &       &       &       &       &       &       \\
		\midrule
		Cosine NN & 0.00  & 0.00  & 0.00  & 0.00  & 0.00  & 0.00  \\
		Cosine SVM & 79.08  & 73.33  & 76.10  & 81.13  & 68.79  & 74.45  \\
		Wavelet NN & 0.00  & 0.00  & 0.00  & 0.00  & 0.00  & 0.00  \\
		Wavelet SVM & 77.53  & 66.95  & 71.85  & 76.05  & 61.13  & 67.78  \\
		Kraw. NN & 50.53  & 15.22  & 23.40  & 50.00  & 15.10  & 23.20  \\
		Kraw. SVM & 50.03  & 65.34  & 56.67  & 49.75  & 48.77  & 49.26  \\
		\midrule
		\emph{Learning:} &       &       &       &       &       &       \\
		\midrule
		SimpleNet & 47.31  & 48.11  & 47.71  & 50.59  & 63.63  & 56.36  \\
		AlexNet & 81.46  & 87.35  & 84.30  & 72.24  & 61.36  & 66.35  \\
		VGGNet & 81.41  & 90.04  & 85.51  & 78.83  & 75.35  & 77.05  \\
		GoogLeNet & 82.74  & 85.46  & 84.08  & 63.35  & 57.74  & 60.42  \\
		ResNet & 80.93  & 84.70  & 82.77  & 68.48  & 66.64  & 67.55  \\
		DenseNet & 87.92  & 93.25  & 90.51  & 82.07  & 83.96  & 83.00  \\
		InceptionNet & 84.60  & 90.92  & 87.65  & 69.58  & 70.77  & 70.17  \\
		MobileNet & 83.07  & 88.07  & 85.50  & 68.73  & 69.50  & 69.11  \\
		
		\midrule
		\emph{Invariant:} &       &       &       &       &       &      \\
		\midrule
		Scatter. NN & 69.85  & 68.94  & 69.39  & 74.93  & 77.31  & 76.10  \\
		Scatter. SVM & 75.70  & 72.07  & 73.84  & 76.42  & 78.63  & 77.51  \\
		HIR NN & 81.27  & 80.68  & 80.98  & 79.09  & 82.17  & 80.60  \\
		HIR SVM & 86.20  & 86.06  & 86.13  & 83.42  & 83.29  & 83.35  \\
		
		\bottomrule
	\end{tabular}
\end{table}%

\subsection{Digital Forensic and Forgery Detection}

For real-world applications, we employ the HIR for large-scale digital forensics, i.e., detections of adversarial perturbation and AIGC, for direct checking its usefulness in robust and interpretable tasks. Note that this plug-and-play strategy will not only be compared to similar representations, but will also a range of current forensic solutions, including well-designed deep forensics.

With the practice of Sections 4, our HIR is implemented here as a single-scale network scale parameter $w = 10$ and composition length $L = 7$; its invariant layer (27) is specialized with $K = {N_{ij}}/2$ and ${\# _B} = 30$, for improving the discriminability of digital artifacts. Note that the feature/architecture selection strategy of Section 4.2 is employed for data adaptability and discriminability, where the top-ranked 500- and 1000-dimensional features are selected for AIGC and adversarial perturbation, respectively. All features are fed into both NN and SVM classifiers, for evaluating the sensitivity w.r.t. the classifier. Unless otherwise stated, the training and testing sets are formed without any crossover by random sampling at 50\% and 50\% ratios on the original dataset, respectively.

\begin{figure}[t]
	\centering
	\includegraphics[scale=0.65]{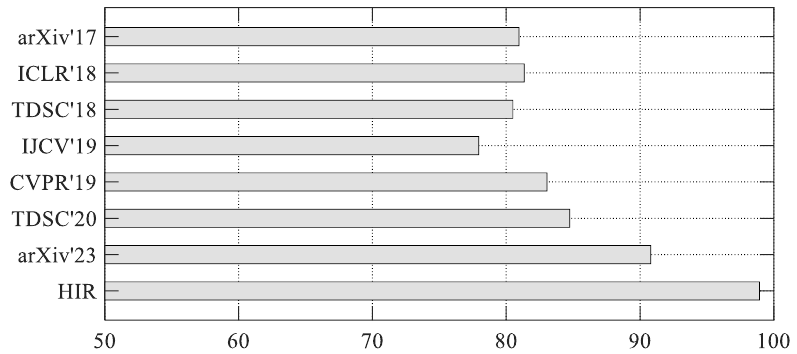}
	\centering
	\caption{A comparison of adversarial perturbation forensic scores (F1, \%) w.r.t. current forensic solutions on the UP benchmark.}
\end{figure}

The competing methods involved here can be summarized as follows:
\begin{itemize}
	\item All the representations in Section 5.1 as direct forensics;
	\item More deep representation milestones for direct forensics, i.e., GoogLeNet \cite{ref58}, ResNet \cite{ref59}, DenseNet \cite{ref60}, InceptionNet \cite{ref61}, and MobileNet \cite{ref62};
	\item The forensic methods designed for adversarial perturbation, i.e., arXiv’17 \cite{ref63}, ICLR’18 \cite{ref64}, TDSC’18 \cite{ref65}, IJCV’19 \cite{ref66}, CVPR’19 \cite{ref67}, TDSC’20 \cite{ref68}, and arXiv’23 \cite{ref69};
	\item The forensic methods designed for AIGC, i.e., ECCV’20 \cite{ref70}, CVPR’20a \cite{ref71}, and CVPR’20b \cite{ref72}.
\end{itemize}

\subsubsection{Adversarial Perturbations}

As shown in Fig. 5, the dataset ImageNet\footnote{https://www.image-net.org/} is perturbed through 6 adversarial methods, i.e., BIM \cite{ref73}, CW \cite{ref74}, Damage \cite{ref75}, FGSM \cite{ref76}, PGD \cite{ref77}, and UP \cite{ref78}, respectively, resulting in 6 benchmarks, each containing 5000 clean images and 5000 perturbed versions. This task exhibits real-world discriminative challenges, in light of the rich variability of the perturbations themselves and the underlying ImageNet.

In Fig. 6, we first provide a comparison with the current solutions of perturbation forensics on the basic and realistic UP benchmark. Despite the fixed perturbation pattern, there are still competing methods failing to achieve good scores. Such methods are with under-complete representations, and thereby unable to comprehensively capture perturbation patterns. In contrast, over-complete arXiv'23 and our HIR all achieve $>$ 90\% scores, further revealing the fundamental role of representation in forensic tasks. Thus, we will next further compare relevant representation strategies.

In Table 5, we train and test all representations on the 6 benchmarks, presenting the corresponding F1 scores, as well as the average and worst score statistics. This protocol exhibits richer intra-class variability over the fixed perturbation.

\begin{itemize}
	\item The frequency difference between natural and perturbed data is a fruitful forensic clue. Therefore, the classical (time)-frequency representations achieve higher scores than generally expected on this large-scale problem. However, such features exhibit significant sensitivity to classifiers. A potential reason is the restricted separability, where one must resort to complex classification strategies in the feature space.
	\item In the learning CNN family, all large-scale networks exhibit $>$ 90\% average scores, especially DenseNet and InceptionNet. The phenomenon suggests that the transfer learning is good at capturing discriminative features with sufficient training data and aligned testing protocol. As for the attacks, the CW is more challenging and dominates the worst scores, mainly due to its variable and weak patterns.
	\item The scattering networks achieve similar scores and much better classifier stability than the original wavelets, suggesting an improvement in the separability. However, its average scores did not reach 90\%, failing to provide a similar level of discriminability as learning CNN.
	\item Our HIR is very robust to classifier changes, also achieving a MobileNet-level of discriminability, slightly lower than DenseNet and InceptionNet, and significantly better than the direct competitor scattering networks. Therefore, our strategy has a better combined performance in robustness, interpretability, and discriminability. Its efficiency benefit will be highlighted in the next experimental protocol.
	
\end{itemize}

In Table 6, we train and test all representations on a hybrid of the 6 perturbation benchmarks, presenting scores at two training-testing ratios. This protocol is more challenging due to very complex intra-class variability, while being more practical for real-world forensic scenarios.

\begin{figure}[t]
	\centering
	\includegraphics[scale=0.65]{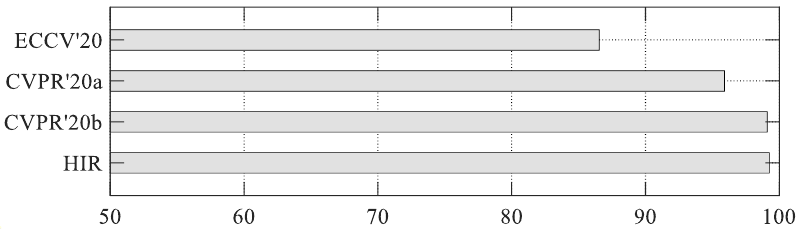}
	\centering
	\caption{A comparison of AIGC forensic scores (F1, \%) w.r.t. current forensic solutions on SD 1.5 benchmark.}
\end{figure}

\begin{table*}[t]
	\centering
	\caption{AIGC Forensic Scores (F1, \%) for Different Representations w.r.t. Various Types of Generators.}
	\begin{tabular}{ccccccccccc}
		\toprule
		Method & ADM & BGAN & GLIDE & Midjourney & SD1.4 & SD1.5 &VQDM & Wukong  & \textbf{Average} & \textbf{Worst} \\
		\midrule
		\emph{Classical:} &       &       &       &       &       &       &       & & &  \\
		\midrule
		Cosine NN & 0.00  & 0.00  & 0.00  & 0.00  & 0.00  & 65.09  & 0.00  & 0.00  & 8.14  & 0.00  \\
		Cosine SVM & 99.19  & 99.95  & 99.57  & 89.02  & 99.10  & 98.80  & 99.46  & 99.11  & 98.03  & 89.02  \\
		Wavelet NN & 0.00  & 0.00  & 0.00  & 0.00  & 0.00  & 2.08  & 0.00  & 0.00  & 0.26  & 0.00  \\
		Wavelet SVM & 99.98  & 99.70  & 99.87  & 85.55  & 98.63  & 99.04  & 99.97  & 99.38  & 97.76  & 85.55  \\
		Krawtchouk NN & 0.00  & 0.00  & 0.00  & 0.00  & 0.00  & 0.00  & 0.00  & 0.00  & 0.00  & 0.00  \\
		Krawtchouk SVM & 99.75  & 99.60  & 98.49  & 62.44  & 74.11  & 77.98  & 93.63  & 76.00  & 85.25  & 62.44  \\
		
		\midrule
		\emph{Learning:} &       &       &       &       &       &       &       & & &  \\
		\midrule
		SimpleNet & 98.25  & 97.87  & 92.98  & 68.00  & 73.52  & 74.37  & 74.88  & 76.32  & 82.02  & 68.00  \\
		AlexNet & 94.45  & 98.99  & 98.26  & 81.52  & 87.96  & 88.66  & 84.24  & 88.62  & 90.34  & 81.52  \\
		VGGNet & 99.40  & 99.38  & 98.57  & 86.44  & 89.97  & 91.86  & 93.60  & 90.09  & 93.66  & 86.44  \\
		GoogLeNet & 80.30  & 99.18  & 98.16  & 75.00  & 82.77  & 82.44  & 86.75  & 82.32  & 85.87  & 75.00  \\
		ResNet & 98.78  & 99.14  & 97.78  & 87.41  & 89.88  & 90.85  & 88.53  & 88.80  & 92.65  & 87.41  \\
		DenseNet & 99.63  & 99.60  & 98.57  & 93.08  & 93.79  & 94.50  & 95.01  & 92.55  & 95.84  & 92.55  \\
		InceptionNet & 97.69  & 99.41  & 98.32  & 90.07  & 89.40  & 92.55  & 92.72  & 88.35  & 93.56  & 88.35  \\
		MobileNet & 90.08  & 99.28  & 97.95  & 87.49  & 88.51  & 90.75  & 87.74  & 88.29  & 91.26  & 87.49  \\
		
		\midrule
		\emph{Invariant:} &       &       &       &       &       &       &       &  & & \\
		\midrule
		ScatterNet NN & 99.10  & 99.63  & 98.44  & 79.47  & 89.26  & 89.95  & 96.71  & 89.07  & 92.70  & 79.47  \\
		ScatterNet SVM & 99.18  & 99.67  & 99.05  & 85.21  & 95.85  & 95.58  & 97.02  & 94.60  & 95.77  & 85.21  \\
		HIR NN & 99.92  & 99.97  & 99.83  & 92.63  & 98.57  & 98.97  & 99.92  & 98.58  & 98.55  & 92.63  \\
		HIR SVM & 99.90  & 99.92  & 99.78  & 92.18  & 99.07  & 99.26  & 99.87  & 99.42  & 98.68  & 92.18  \\
		
		\bottomrule
	\end{tabular}%
	
\end{table*}%

\begin{itemize}
	\item In line with previous observations, the classical representations still exhibit score fluctuations on the two classifiers. We also note a performance degradation compared to the case of Table 5, due to the discriminative challenges by this hybrid protocol. On the other hand, their performance is stable w.r.t. the reduction of training samples, further validating the inherent advantages in sample efficiency.
	\item Moving into this hybrid benchmark, the learning CNN family yields consistent and large performance degradation, especially for the $1/9$ case with fewer samples. This phenomenon is direct evidence for the data dependence in learning representations (even with transfer strategy). In fact, real-world forensics often face the situation where the perturbation types are diverse and some of them lack samples. Therefore, such data-dependent forensics typically exhibit time-consuming (re-)training, while failing to guarantee their validity for under-sampled perturbation patterns.
	\item The scattering networks basically continue the discriminability level and classifier stability from Table 5. Note that its scores in the $1/9$ case are higher than most classical and learning representations, reflecting the superior performance in both discriminability and efficiency.
	\item In this challenging protocol, the hand-crafted HIR still achieves a learning-level discriminability and consistently outperforms scattering networks. More importantly, our HIR is significantly less dependent on training samples than learning CNN, meaning it can better cope with under-sampled perturbation patterns in practice. For the next larger-scale forensic task, the comprehensive advantages of HIR over learning CNN will be further highlighted, in robustness, interpretability, discriminability, and efficiency.
	
\end{itemize}

\begin{table*}[t]
	\centering
	\caption{AIGC Forensic Robustness Scores (F1, \%) for Different Representations w.r.t. Various Types of Generators.}
	\begin{tabular}{ccccccccccc}
		\toprule
		\multirow{2}{*}{Method} & \multicolumn{10}{c}{\textbf{Testing With Random Orientation and Flipping}} \\
		& ADM & BGAN & GLIDE & Midjourney & SD1.4 & SD1.5 &VQDM & Wukong  & \textbf{Average} & \textbf{Worst} \\
		\midrule
		\emph{Classical:} &       &       &       &       &       &       &       & & &  \\
		\midrule
		Cosine SVM & 99.16  & 99.95  & 99.55  & 88.07  & 99.05  & 98.72  & 99.36  & 99.11  & 97.87  & 88.07  \\
		Wavelet SVM & 99.95  & 99.80  & 99.85  & 82.83  & 99.08  & 98.71  & 99.95  & 99.16  & 97.42  & 82.83  \\
		Krawtchouk SVM & 70.90  & 99.58  & 98.57  & 64.90  & 77.28  & 76.58  & 94.44  & 76.72  & 82.37  & 64.90  \\
		
		\midrule
		\emph{Learning:} &       &       &       &       &       &       &       & & &  \\
		\midrule
		SimpleNet & 77.72  & 95.04  & 92.99  & 65.25  & 74.52  & 74.72  & 73.62  & 76.91  & 78.85  & 65.25  \\
		AlexNet & 81.82  & 99.08  & 97.99  & 77.45  & 85.87  & 87.93  & 83.29  & 86.78  & 87.53  & 77.45  \\
		VGGNet & 76.18  & 99.40  & 98.41  & 82.23  & 89.26  & 89.30  & 88.79  & 89.00  & 89.07  & 76.18  \\
		GoogLeNet & 80.62  & 99.30  & 98.09  & 73.13  & 81.29  & 81.92  & 85.36  & 82.31  & 85.25  & 73.13  \\
		ResNet & 85.62  & 99.28  & 97.60  & 81.82  & 85.60  & 86.84  & 87.94  & 85.21  & 88.74  & 81.82  \\
		DenseNet & 84.57  & 99.56  & 98.66  & 88.57  & 91.16  & 91.60  & 94.26  & 89.47  & 92.23  & 84.57  \\
		InceptionNet & 91.99  & 99.30  & 98.56  & 87.04  & 85.72  & 88.85  & 92.19  & 85.24  & 91.11  & 85.24  \\
		MobileNet & 85.14  & 99.36  & 97.78  & 84.54  & 86.74  & 88.80  & 87.50  & 86.56  & 89.55  & 84.54  \\
		
		\midrule
		\emph{Invariant:} &       &       &       &       &       &       &       &  & & \\
		\midrule
		ScatterNet SVM & 92.61  & 99.67  & 99.18  & 82.75  & 88.81  & 89.40  & 97.05  & 83.83  & 91.66  & 82.75  \\
		HIR SVM & 99.88  & 99.87  & 99.77  & 91.99  & 99.02  & 99.25  & 99.83  & 99.45  & 98.63  & 91.99  \\
		
		\bottomrule
	\end{tabular}%
	
\end{table*}%

\begin{table}[t]
	\centering
	\caption{AIGC Forensic Scores (\%) for Different Representations on a Real-world (Hybrid) Benchmark}
	\begin{tabular}{ccccccc}
		\toprule
		\multirow{2}{*}{Method} & \multicolumn{3}{c}{\textbf{Train./Test. = 5/5}} & \multicolumn{3}{c}{\textbf{Train./Test. = 1/9}} \\
		&   Pre. & Rec. & F1 & Pre. & Rec. & F1 \\
		\midrule
		\emph{Classical:} &       &       &       &       &       &       \\
		\midrule
		Cosine NN & 0.00  & 0.00  & 0.00  & 0.00  & 0.00  & 0.00  \\
		Cosine SVM & 94.95  & 94.57  & 94.76  & 94.36  & 91.06  & 92.68  \\
		Wavelet NN & 48.70  & 94.17  & 64.20  & 48.69  & 94.13  & 64.18  \\
		Wavelet SVM & 94.03  & 94.57  & 94.30  & 83.55  & 93.48  & 88.24  \\
		Kraw. NN & 0.00  & 0.00  & 0.00  & 0.00  & 0.00  & 0.00  \\
		Kraw. SVM & 75.24  & 74.77  & 75.00  & 71.56  & 68.57  & 70.03  \\
		
		\midrule
		\emph{Learning:} &       &       &       &       &       &       \\
		\midrule
		SimpleNet & 61.79  & 40.70  & 49.08  & 56.40  & 60.48  & 58.37  \\
		AlexNet & 80.76  & 77.63  & 79.16  & 71.83  & 72.50  & 72.17  \\
		VGGNet & 84.75  & 86.67  & 85.70  & 72.45  & 72.37  & 72.41  \\
		GoogLeNet & 74.15  & 80.40  & 77.15  & 67.84  & 68.83  & 68.33  \\
		ResNet & 85.10  & 83.03  & 84.06  & 76.88  & 73.67  & 75.24  \\
		DenseNet & 86.83  & 85.23  & 86.02  & 76.84  & 75.37  & 76.10  \\
		InceptionNet & 82.69  & 86.63  & 84.62  & 68.62  & 68.56  & 68.59  \\
		MobileNet & 81.54  & 82.47  & 82.00  & 68.52  & 68.57  & 68.55  \\
		
		\midrule
		\emph{Invariant:} &       &       &       &       &       &      \\
		\midrule
		Scatter. NN & 83.68  & 83.73  & 83.71  & 79.37  & 79.70  & 79.53  \\
		Scatter. SVM & 90.31  & 85.17  & 87.67  & 85.28  & 79.70  & 82.40  \\
		HIR NN & 96.79  & 96.47  & 96.63  & 95.66  & 93.04  & 94.33  \\
		HIR SVM & 96.92  & 96.37  & 96.64  & 95.21  & 94.26  & 94.73  \\
		
		\bottomrule
	\end{tabular}
\end{table}%

\subsubsection{Artificial Intelligence Generated Content}

As shown in Fig. 5, fake images with similar content to ImageNet are synthesized through 8 AIGC methods, i.e., ADM \cite{ref79}, BGAN \cite{ref80}, GLIDE \cite{ref81}, Midjourney\footnote{https://www.midjourney.com/home}, SD 1.4 \cite{ref82}, SD 1.5 \cite{ref82}, VQDM \cite{ref83}, and Wukong\footnote{https://xihe.mindspore.cn/modelzoo/wukong}, respectively, resulting in 8 benchmarks, each containing 6000 natural images and 6000 synthesized images. This task exhibits higher level of discriminative challenges, in light of the very rich variability of both natural and synthesized content.

In Fig. 7, we first provide a comparison with the current solutions of AIGC forensics on the basic SD 1.5 benchmark. All these methods based on deep networks and feature enhancements achieve good forensic scores. This indicates that the forensic scenario with sufficient training and aligned testing is not challenging for typical learning representations. Here, the hand-crafted HIR also achieves $\sim$ 100\% scores, meaning a similar discriminability for this forensic scenario. Next, we will compare scores in a comprehensive manner, mainly at the representation level, and also some AIGC forensic solutions as references.

In Table 7, we train and test all representations on the 8 benchmarks, presenting the corresponding F1 scores, as well as the average and worst score statistics. Besides this direct protocol, we also consider testing image variants with random orientation or flipping in Table 8, reflecting the basic geometric robustness requirements.

\begin{itemize}
	\item One can observe that the frequency forensic clue of the AI-generated pipeline is still very effective. The classical representations based on SVM classifier achieve consistently good accuracy and robustness, suggesting that state-of-the-art generators (even diffusion ones) still exhibit inherent frequency artifacts. On the other hand, such hand-crafted features are sensitive to classifiers, in line with the observations of Section 5.2.1.
	\item The learning representations other than SimpleNet and GoogLeNet achieve $>$ 90\% average scores, further confirming their good discriminability with sufficient training data and aligned testing protocol. However, for the robustness protocol in Table 8, they exhibit varying degrees of performance degradation, in both average and worst statistics. Clearly, even natural and slight shifts in the data distribution can strongly interfere with the learning forensics. In particular, such interference is highly black-boxed (i.e., unpredictable), where an example is the significantly higher fluctuations on ADM compared to others.
	\item The scattering networks exhibit similar level of discriminability and robustness as the learning representations, while outperforming the original wavelets in classifier stability. Note that scattering networks fails to achieve the expected invariance and thus cannot provide higher robustness scores than learning CNN.
	\item Regarding the discriminability, geometric invariance, and classifier stability, our HIR achieves better combined performance versus classical representations, scattering networks, and learning CNN. This is in line with our theory expectation that HIR combines the advantages of both hand-crafted and learning representations. While its efficiency will be further highlighted in the next experimental protocol.
	
\end{itemize}

In Table 9, we train and test all representations on a hybrid of the 8 AIGC benchmarks, presenting scores at two training-testing ratios. This protocol is more challenging due to very complex intra-class variability, while being more practical for real-world forensic scenarios.

\begin{itemize}
	\item The classical representations exhibit good discriminability for this hybrid benchmark, along with the benefit of sample efficiency. In line with previous observations, they still exhibit score fluctuations on the two classifiers.
	\item The forensic scores of the learning CNN family drop significantly, with average scores of only $\sim$70\% in the $1/9$ case. This further illustrates the weakness of learning forensic algorithms in dealing with real-world scenarios, i.e., the data dependence problem.
	\item The scattering networks inherits the discriminability level and classifier stability from Table 7, also with better scores than all learning representations in the $1/9$ case. These all indicate the superiority in combined performance.
	\item Here, the HIR achieves the highest scores over classical representations, scattering networks, and learning CNN. Its discriminability allows for a hybrid forensic of the 8 AIGC methods in the $1/9$ case with fewer samples, implying the usefulness in real-world forensic scenarios. Note that the ECCV’20, CVPR’20a, and CVPR’20b specifically designed for AIGC forensics exhibit 79.67\%, 73.74\%, and 76.89\% F1 scores in the $1/9$ case, respectively. In summary, our HIR yields consistently good discriminability, robustness, and efficiency in a plug-and-play way, spanning 3 classical vision tasks and 2 forensic tasks. No representation and forensic method achieves such results in these experiments.
\end{itemize}

\section{Conclusion}

In this paper, we have systematically investigated the topic of hierarchical invariance, as an early attempt to harmonize the divergence between typical CNN and invariants w.r.t. discriminability and robustness. Compared to related research approaches, our hierarchical invariant representation can be characterized as 1) principled and interpretable design, 2) efficient invariant structure, and 3) competitive discriminability in the era of deep learning.

The \emph{theory} ingredients of our work are as follows.

\begin{itemize}
	\item We have formalized a blueprint for hierarchical invariance, rethinking the typical modules of CNN representations.
	\item We have defined new modules with their compositions to fulfill the blueprint, providing formal conclusions about the geometric symmetries between image and representation.
	\item We have discussed the criticisms and developments of the above new idea versus typically concepts, highlighting our uniqueness in moving towards robust and interpretable representations.
	
\end{itemize}

The \emph{practice} ingredients of our work are as follows.

\begin{itemize}
	\item We have provided a specific framework for the theory of hierarchical invariance, covering practical principles about the topologies, layers, and parameters.
	\item We have explored the data adaptability potential of the above framework, resorting to feature/architecture selection or cascading learning module.
	
\end{itemize}

The \emph{application} ingredients of our work are as follows.

\begin{itemize}
	\item We have conducted pattern classification experiments on typical sets of texture, digit, and parasite images, respectively, examining the properties promised by our theory under diverse task scales and geometric variants.
	\item We have checked the realistic usefulness in large-scale digital forensics of adversarial perturbation and AIGC. Here, the HIR yields consistently good discriminability, robustness, efficiency, and interpretability in a plug-and-play way, exhibiting competitive overall performance than current representation and forensic methods.

\end{itemize}

\ifCLASSOPTIONcaptionsoff
  \newpage
\fi



%

\bibliographystyle{IEEEtran}
\bibliography{paper}

\begin{thebibliography}{10}
\providecommand{\url}[1]{#1}
\csname url@samestyle\endcsname
\providecommand{\newblock}{\relax}
\providecommand{\bibinfo}[2]{#2}
\providecommand{\BIBentrySTDinterwordspacing}{\spaceskip=0pt\relax}
\providecommand{\BIBentryALTinterwordstretchfactor}{4}
\providecommand{\BIBentryALTinterwordspacing}{\spaceskip=\fontdimen2\font plus
\BIBentryALTinterwordstretchfactor\fontdimen3\font minus
  \fontdimen4\font\relax}
\providecommand{\BIBforeignlanguage}[2]{{%
\expandafter\ifx\csname l@#1\endcsname\relax
\typeout{** WARNING: IEEEtran.bst: No hyphenation pattern has been}%
\typeout{** loaded for the language `#1'. Using the pattern for}%
\typeout{** the default language instead.}%
\else
\language=\csname l@#1\endcsname
\fi
#2}}
\providecommand{\BIBdecl}{\relax}
\BIBdecl

\bibitem{ref1}
J.~M. Wing, ``Trustworthy {AI},'' \emph{Commun. ACM}, vol.~64, no.~10, pp.
  64--71, 2021.

\bibitem{ref2}
Y.~LeCun, Y.~Bengio, and G.~Hinton, ``Deep learning,'' \emph{Nature}, vol. 521,
  no. 7553, pp. 436--444, 2015.

\bibitem{ref3}
D.~Silver, J.~Schrittwieser, K.~Simonyan, I.~Antonoglou, A.~Huang, A.~Guez,
  T.~Hubert, L.~Baker, M.~Lai, A.~Bolton \emph{et~al.}, ``Mastering the game of
  go without human knowledge,'' \emph{Nature}, vol. 550, no. 7676, pp.
  354--359, 2017.

\bibitem{ref4}
H.~Wang, T.~Fu, Y.~Du, W.~Gao, K.~Huang, Z.~Liu, P.~Chandak, S.~Liu,
  P.~Van~Katwyk, A.~Deac \emph{et~al.}, ``Scientific discovery in the age of
  artificial intelligence,'' \emph{Nature}, vol. 620, no. 7972, pp. 47--60,
  2023.

\bibitem{ref5}
K.~Sundararajan and D.~L. Woodard, ``Deep learning for biometrics: A survey,''
  \emph{ACM Comput. Surv.}, vol.~51, no.~3, pp. 1--34, 2018.

\bibitem{ref6}
S.~Warnat-Herresthal, H.~Schultze, K.~L. Shastry, S.~Manamohan, S.~Mukherjee,
  V.~Garg, R.~Sarveswara, K.~H{\"a}ndler, P.~Pickkers, N.~A. Aziz
  \emph{et~al.}, ``Swarm learning for decentralized and confidential clinical
  machine learning,'' \emph{Nature}, vol. 594, no. 7862, pp. 265--270, 2021.

\bibitem{ref7}
S.~Feng, H.~Sun, X.~Yan, H.~Zhu, Z.~Zou, S.~Shen, and H.~X. Liu, ``Dense
  reinforcement learning for safety validation of autonomous vehicles,''
  \emph{Nature}, vol. 615, no. 7953, pp. 620--627, 2023.

\bibitem{ref8}
F.~Juefei-Xu, R.~Wang, Y.~Huang, Q.~Guo, L.~Ma, and Y.~Liu, ``Countering
  malicious deepfakes: Survey, battleground, and horizon,'' \emph{Int. J.
  Comput. Vis.}, vol. 130, no.~7, pp. 1678--1734, 2022.

\bibitem{ref9}
H.~Liu, M.~Chaudhary, and H.~Wang, ``Towards trustworthy and aligned machine
  learning: A data-centric survey with causality perspectives,'' \emph{arXiv
  preprint arXiv:2307.16851}, 2023.

\bibitem{ref10}
M.~M. Bronstein, J.~Bruna, T.~Cohen, and P.~Veli{\v{c}}kovi{\'c}, ``Geometric
  deep learning: Grids, groups, graphs, geodesics, and gauges,'' \emph{arXiv
  preprint arXiv:2104.13478}, 2021.

\bibitem{ref11}
Y.~Bengio, A.~Courville, and P.~Vincent, ``Representation learning: A review
  and new perspectives,'' \emph{IEEE Trans. Pattern Anal. Mach. Intell.},
  vol.~35, no.~8, pp. 1798--1828, 2013.

\bibitem{ref12}
D.~Ulyanov, A.~Vedaldi, and V.~Lempitsky, ``Deep image prior,'' in \emph{Proc.
  IEEE Conf. Comput. Vis. Pattern Recognit.}, 2018, pp. 9446--9454.

\bibitem{ref13}
F.~Klein, ``A comparative review of recent researches in geometry,''
  \emph{Bull. Am. Math. Soc.}, vol.~2, no.~10, pp. 215--249, 1893.

\bibitem{ref14}
R.~Tobies, \emph{{Felix Klein}---Mathematician, Academic Organizer, Educational
  Reformer}.\hskip 1em plus 0.5em minus 0.4em\relax Cham: Springer
  International Publishing, 2019, pp. 5--21.

\bibitem{ref15}
K.~Fukushima and S.~Miyake, ``Neocognitron: A new algorithm for pattern
  recognition tolerant of deformations and shifts in position,'' \emph{Pattern
  Recognit.}, vol.~15, no.~6, pp. 455--469, 1982.

\bibitem{ref16}
V.~Balntas, K.~Lenc, A.~Vedaldi, T.~Tuytelaars, J.~Matas, and K.~Mikolajczyk,
  ``{H-Patches}: A benchmark and evaluation of handcrafted and learned local
  descriptors.'' \emph{IEEE Trans. Pattern Anal. Mach. Intell.}, vol.~42,
  no.~11, pp. 2825--2841, 2019.

\bibitem{ref17}
S.~Qi, Y.~Zhang, C.~Wang, J.~Zhou, and X.~Cao, ``A survey of orthogonal moments
  for image representation: theory, implementation, and evaluation,'' \emph{ACM
  Comput. Surv.}, vol.~55, no.~1, pp. 1--35, 2021.

\bibitem{ref18}
D.~G. Lowe, ``Distinctive image features from scale-invariant keypoints,''
  \emph{Int. J. Comput. Vis.}, vol.~60, pp. 91--110, 2004.

\bibitem{ref19}
E.~Tola, V.~Lepetit, and P.~Fua, ``Daisy: An efficient dense descriptor applied
  to wide-baseline stereo,'' \emph{IEEE Trans. Pattern Anal. Mach. Intell.},
  vol.~32, no.~5, pp. 815--830, 2009.

\bibitem{ref20}
O.~Russakovsky, J.~Deng, H.~Su, J.~Krause, S.~Satheesh, S.~Ma, Z.~Huang,
  A.~Karpathy, A.~Khosla, M.~Bernstein \emph{et~al.}, ``{ImageNet} large scale
  visual recognition challenge,'' \emph{Int. J. Comput. Vis.}, vol. 115, pp.
  211--252, 2015.

\bibitem{ref21}
R.~Zhang, ``Making convolutional networks shift-invariant again,'' in
  \emph{Proc. Int. Conf. Mach. Learn.}, 2019, pp. 7324--7334.

\bibitem{ref22}
C.~Buckner, ``Understanding adversarial examples requires a theory of artefacts
  for deep learning,'' \emph{Nature Mach. Intell.}, vol.~2, no.~12, pp.
  731--736, 2020.

\bibitem{ref23}
M.~Taddeo, T.~McCutcheon, and L.~Floridi, ``Trusting artificial intelligence in
  cybersecurity is a double-edged sword,'' \emph{Nature Mach. Intell.}, vol.~1,
  no.~12, pp. 557--560, 2019.

\bibitem{ref24}
J.~Bruna and S.~Mallat, ``Invariant scattering convolution networks,''
  \emph{IEEE Trans. Pattern Anal. Mach. Intell.}, vol.~35, no.~8, pp.
  1872--1886, 2013.

\bibitem{ref25}
L.~Sifre and S.~Mallat, ``Rotation, scaling and deformation invariant
  scattering for texture discrimination,'' in \emph{Proc. IEEE Conf. Comput.
  Vis. Pattern Recognit.}, 2013, pp. 1233--1240.

\bibitem{ref26}
T.~Wiatowski and H.~B{\"o}lcskei, ``A mathematical theory of deep convolutional
  neural networks for feature extraction,'' \emph{IEEE Trans. Inf. Theory},
  vol.~64, no.~3, pp. 1845--1866, 2017.

\bibitem{ref27}
E.~Oyallon, S.~Zagoruyko, G.~Huang, N.~Komodakis, S.~Lacoste-Julien,
  M.~Blaschko, and E.~Belilovsky, ``Scattering networks for hybrid
  representation learning,'' \emph{IEEE Trans. Pattern Anal. Mach. Intell.},
  vol.~41, no.~9, pp. 2208--2221, 2018.

\bibitem{ref28}
J.~And{\'e}n and S.~Mallat, ``Deep scattering spectrum,'' \emph{IEEE Trans.
  Signal Process.}, vol.~62, no.~16, pp. 4114--4128, 2014.

\bibitem{ref29}
X.~Chen, X.~Cheng, and S.~Mallat, ``Unsupervised deep {Haar} scattering on
  graphs,'' \emph{Proc. Adv. Neural Inf. Process. Syst.}, vol.~27, 2014.

\bibitem{ref30}
S.~Yu, ``Evolving scattering networks for engineering disorder,'' \emph{Nature
  Comput. Sci.}, vol.~3, no.~2, pp. 128--138, 2023.

\bibitem{ref31}
S.~Cheng, Y.-S. Ting, B.~M{\'e}nard, and J.~Bruna, ``A new approach to
  observational cosmology using the scattering transform,'' \emph{Mon. Not. R.
  Astron. Soc.}, vol. 499, no.~4, pp. 5902--5914, 2020.

\bibitem{ref32}
T.~Cohen and M.~Welling, ``Group equivariant convolutional networks,'' in
  \emph{Proc. Int. Conf. Mach. Learn.}, 2016, pp. 2990--2999.

\bibitem{ref33}
T.~S. Cohen and M.~Welling, ``Steerable {CNNs},'' in \emph{Proc. Int. Conf.
  Learn. Representations}, 2016.

\bibitem{ref34}
D.~E. Worrall, S.~J. Garbin, D.~Turmukhambetov, and G.~J. Brostow, ``Harmonic
  networks: Deep translation and rotation equivariance,'' in \emph{Proc. IEEE
  Conf. Comput. Vis. Pattern Recognit.}, 2017, pp. 5028--5037.

\bibitem{ref35}
M.~Weiler, F.~A. Hamprecht, and M.~Storath, ``Learning steerable filters for
  rotation equivariant {CNNs},'' in \emph{Proc. IEEE Conf. Comput. Vis. Pattern
  Recognit.}, 2018, pp. 849--858.

\bibitem{ref36}
I.~Sosnovik, M.~Szmaja, and A.~Smeulders, ``Scale-equivariant steerable
  networks,'' in \emph{Proc. Int. Conf. Learn. Representations}, 2019.

\bibitem{ref37}
D.~Worrall and M.~Welling, ``Deep scale-spaces: Equivariance over scale,''
  \emph{Proc. Adv. Neural Inf. Process. Syst.}, vol.~32, 2019.

\bibitem{ref38}
Z.~Sun and T.~Blu, ``Empowering networks with scale and rotation equivariance
  using a similarity convolution,'' in \emph{Proc. Int. Conf. Learn.
  Representations}, 2022.

\bibitem{ref39}
M.~Finzi, S.~Stanton, P.~Izmailov, and A.~G. Wilson, ``Generalizing
  convolutional neural networks for equivariance to {Lie} groups on arbitrary
  continuous data,'' in \emph{Proc. Int. Conf. Mach. Learn.}, 2020, pp.
  3165--3176.

\bibitem{ref40}
E.~J. Bekkers, ``B-spline {CNNs} on {Lie} groups,'' in \emph{Proc. Int. Conf.
  Learn. Representations}, 2019.

\bibitem{ref41}
Q.~Xie, Q.~Zhao, Z.~Xu, and D.~Meng, ``Fourier series expansion based filter
  parametrization for equivariant convolutions,'' \emph{IEEE Trans. Pattern
  Anal. Mach. Intell.}, vol.~45, no.~4, pp. 4537--4551, 2022.

\bibitem{ref42}
K.~Atz, F.~Grisoni, and G.~Schneider, ``Geometric deep learning on molecular
  representations,'' \emph{Nature Mach. Intell.}, vol.~3, no.~12, pp.
  1023--1032, 2021.

\bibitem{ref43}
R.~J. Townshend, S.~Eismann, A.~M. Watkins, R.~Rangan, M.~Karelina, R.~Das, and
  R.~O. Dror, ``Geometric deep learning of {RNA} structure,'' \emph{Science},
  vol. 373, no. 6558, pp. 1047--1051, 2021.

\bibitem{ref44}
I.~Goodfellow, P.~McDaniel, and N.~Papernot, ``Making machine learning robust
  against adversarial inputs,'' \emph{Commun. ACM}, vol.~61, no.~7, pp. 56--66,
  2018.

\bibitem{ref45}
F.~Zhan, Y.~Yu, R.~Wu, J.~Zhang, S.~Lu, L.~Liu, A.~Kortylewski, C.~Theobalt,
  and E.~Xing, ``Multimodal image synthesis and editing: The generative {AI}
  era,'' \emph{IEEE Trans. Pattern Anal. Mach. Intell.}, 2023.

\bibitem{ref46}
S.~Qi, Y.~Zhang, C.~Wang, J.~Zhou, and X.~Cao, ``A principled design of image
  representation: Towards forensic tasks,'' \emph{IEEE Trans. Pattern Anal.
  Mach. Intell.}, vol.~45, no.~5, pp. 5337--5354, 2022.

\bibitem{ref47}
K.~Lenc and A.~Vedaldi, ``Understanding image representations by measuring
  their equivariance and equivalence,'' in \emph{Proc. IEEE Conf. Comput. Vis.
  Pattern Recognit.}, 2015, pp. 991--999.

\bibitem{ref48}
D.~Marcos, M.~Volpi, N.~Komodakis, and D.~Tuia, ``Rotation equivariant vector
  field networks,'' in \emph{Proc. IEEE Int. Conf. Comput. Vis.}, 2017, pp.
  5048--5057.

\bibitem{ref49}
S.~Qi, Y.~Zhang, C.~Wang, T.~Xiang, X.~Cao, and Y.~Xiang, ``Representing noisy
  image without denoising,'' \emph{arXiv preprint arXiv:2301.07409}, 2023.

\bibitem{ref50}
P.-T. Yap, X.~Jiang, and A.~C. Kot, ``Two-dimensional polar harmonic transforms
  for invariant image representation,'' \emph{IEEE Trans. Pattern Anal. Mach.
  Intell.}, vol.~32, no.~7, pp. 1259--1270, 2009.

\bibitem{ref51}
J.~Flusser, B.~Zitova, and T.~Suk, \emph{Moments and moment invariants in
  pattern recognition}.\hskip 1em plus 0.5em minus 0.4em\relax John Wiley \&
  Sons, 2009.

\bibitem{ref52}
G.~Bender, P.-J. Kindermans, B.~Zoph, V.~Vasudevan, and Q.~Le, ``Understanding
  and simplifying one-shot architecture search,'' in \emph{Proc. Int. Conf.
  Mach. Learn.}, 2018, pp. 550--559.

\bibitem{ref53}
M.~Guo, Y.~Yang, R.~Xu, Z.~Liu, and D.~Lin, ``When {NAS} meets robustness: In
  search of robust architectures against adversarial attacks,'' in \emph{Proc.
  IEEE Conf. Comput. Vis. Pattern Recognit.}, 2020, pp. 631--640.

\bibitem{ref54}
S.~G. Mallat, ``A theory for multiresolution signal decomposition: The wavelet
  representation,'' \emph{IEEE Trans. Pattern Anal. Mach. Intell.}, vol.~11,
  no.~7, pp. 674--693, 1989.

\bibitem{ref55}
P.-T. Yap, R.~Paramesran, and S.-H. Ong, ``Image analysis by {Krawtchouk}
  moments,'' \emph{IEEE Trans. Image Process.}, vol.~12, no.~11, pp.
  1367--1377, 2003.

\bibitem{ref56}
A.~Krizhevsky, I.~Sutskever, and G.~E. Hinton, ``{ImageNet} classification with
  deep convolutional neural networks,'' \emph{Commun. ACM}, vol.~60, no.~6, pp.
  84--90, 2017.

\bibitem{ref57}
K.~Simonyan and A.~Zisserman, ``Very deep convolutional networks for
  large-scale image recognition,'' \emph{arXiv preprint arXiv:1409.1556}, 2014.

\bibitem{ref58}
C.~Szegedy, W.~Liu, Y.~Jia, P.~Sermanet, S.~Reed, D.~Anguelov, D.~Erhan,
  V.~Vanhoucke, and A.~Rabinovich, ``Going deeper with convolutions,'' in
  \emph{Proc. IEEE Conf. Comput. Vis. Pattern Recognit.}, 2015, pp. 1--9.

\bibitem{ref59}
K.~He, X.~Zhang, S.~Ren, and J.~Sun, ``Deep residual learning for image
  recognition,'' in \emph{Proc. IEEE Conf. Comput. Vis. Pattern Recognit.},
  2016, pp. 770--778.

\bibitem{ref60}
G.~Huang, Z.~Liu, L.~Van Der~Maaten, and K.~Q. Weinberger, ``Densely connected
  convolutional networks,'' in \emph{Proc. IEEE Conf. Comput. Vis. Pattern
  Recognit.}, 2017, pp. 4700--4708.

\bibitem{ref61}
C.~Szegedy, V.~Vanhoucke, S.~Ioffe, J.~Shlens, and Z.~Wojna, ``Rethinking the
  inception architecture for computer vision,'' in \emph{Proc. IEEE Conf.
  Comput. Vis. Pattern Recognit.}, 2016, pp. 2818--2826.

\bibitem{ref62}
M.~Sandler, A.~Howard, M.~Zhu, A.~Zhmoginov, and L.-C. Chen, ``Mobilenet-v2:
  Inverted residuals and linear bottlenecks,'' in \emph{Proc. IEEE Conf.
  Comput. Vis. Pattern Recognit.}, 2018, pp. 4510--4520.

\bibitem{ref63}
R.~Feinman, R.~R. Curtin, S.~Shintre, and A.~B. Gardner, ``Detecting
  adversarial samples from artifacts,'' \emph{arXiv preprint arXiv:1703.00410},
  2017.

\bibitem{ref64}
S.~Liang, Y.~Li, and R.~Srikant, ``Enhancing the reliability of
  out-of-distribution image detection in neural networks,'' in \emph{Proc. Int.
  Conf. Learn. Representations}, 2018.

\bibitem{ref65}
B.~Liang, H.~Li, M.~Su, X.~Li, W.~Shi, and X.~Wang, ``Detecting adversarial
  image examples in deep neural networks with adaptive noise reduction,''
  \emph{IEEE Trans. Dependable Secure Comput.}, vol.~18, no.~1, pp. 72--85,
  2018.

\bibitem{ref66}
G.~Goswami, A.~Agarwal, N.~Ratha, R.~Singh, and M.~Vatsa, ``Detecting and
  mitigating adversarial perturbations for robust face recognition,''
  \emph{Int. J. Comput. Vis.}, vol. 127, pp. 719--742, 2019.

\bibitem{ref67}
J.~Liu, W.~Zhang, Y.~Zhang, D.~Hou, Y.~Liu, H.~Zha, and N.~Yu, ``Detection
  based defense against adversarial examples from the steganalysis point of
  view,'' in \emph{Proc. IEEE Conf. Comput. Vis. Pattern Recognit.}, 2019, pp.
  4825--4834.

\bibitem{ref68}
A.~Agarwal, R.~Singh, M.~Vatsa, and N.~Ratha, ``Image transformation-based
  defense against adversarial perturbation on deep learning models,''
  \emph{IEEE Trans. Dependable Secure Comput.}, vol.~18, no.~5, pp. 2106--2121,
  2020.

\bibitem{ref69}
C.~Wang, S.~Qi, Z.~Huang, Y.~Zhang, R.~Lan, and X.~Cao, ``Towards an accurate
  and secure detector against adversarial perturbations,'' \emph{arXiv preprint
  arXiv:2305.10856}, 2023.

\bibitem{ref70}
Y.~Qian, G.~Yin, L.~Sheng, Z.~Chen, and J.~Shao, ``Thinking in frequency: Face
  forgery detection by mining frequency-aware clues,'' in \emph{Proc. Eur.
  Conf. Comput. Vis.}, 2020, pp. 86--103.

\bibitem{ref71}
S.-Y. Wang, O.~Wang, R.~Zhang, A.~Owens, and A.~A. Efros, ``{CNN}-generated
  images are surprisingly easy to spot... for now,'' in \emph{Proc. IEEE Conf.
  Comput. Vis. Pattern Recognit.}, 2020, pp. 8695--8704.

\bibitem{ref72}
Z.~Liu, X.~Qi, and P.~H. Torr, ``Global texture enhancement for fake face
  detection in the wild,'' in \emph{Proc. IEEE Conf. Comput. Vis. Pattern
  Recognit.}, 2020, pp. 8060--8069.

\bibitem{ref73}
A.~Kurakin, I.~J. Goodfellow, and S.~Bengio, ``Adversarial examples in the
  physical world,'' in \emph{Artificial Intelligence Safety and
  Security}.\hskip 1em plus 0.5em minus 0.4em\relax Chapman and Hall/CRC, 2018,
  pp. 99--112.

\bibitem{ref74}
N.~Carlini and D.~Wagner, ``Towards evaluating the robustness of neural
  networks,'' in \emph{Proc. IEEE Symp. Secur. Privacy}, 2017, pp. 39--57.

\bibitem{ref75}
S.~Chen, Z.~He, C.~Sun, J.~Yang, and X.~Huang, ``Universal adversarial attack
  on attention and the resulting dataset {DamageNet},'' \emph{IEEE Trans.
  Pattern Anal. Mach. Intell.}, vol.~44, no.~4, pp. 2188--2197, 2020.

\bibitem{ref76}
I.~J. Goodfellow, J.~Shlens, and C.~Szegedy, ``Explaining and harnessing
  adversarial examples,'' \emph{arXiv preprint arXiv:1412.6572}, 2014.

\bibitem{ref77}
A.~Madry, A.~Makelov, L.~Schmidt, D.~Tsipras, and A.~Vladu, ``Towards deep
  learning models resistant to adversarial attacks,'' in \emph{Proc. Int. Conf.
  Learn. Representations}, 2018.

\bibitem{ref78}
S.-M. Moosavi-Dezfooli, A.~Fawzi, O.~Fawzi, and P.~Frossard, ``Universal
  adversarial perturbations,'' in \emph{Proc. IEEE Conf. Comput. Vis. Pattern
  Recognit.}, 2017, pp. 1765--1773.

\bibitem{ref79}
P.~Dhariwal and A.~Nichol, ``Diffusion models beat {GANs} on image synthesis,''
  \emph{Proc. Adv. Neural Inf. Process. Syst.}, vol.~34, pp. 8780--8794, 2021.

\bibitem{ref80}
A.~Brock, J.~Donahue, and K.~Simonyan, ``Large scale {GAN} training for high
  fidelity natural image synthesis,'' \emph{arXiv preprint arXiv:1809.11096},
  2018.

\bibitem{ref81}
A.~Nichol, P.~Dhariwal, A.~Ramesh, P.~Shyam, P.~Mishkin, B.~McGrew,
  I.~Sutskever, and M.~Chen, ``Glide: Towards photorealistic image generation
  and editing with text-guided diffusion models,'' \emph{arXiv preprint
  arXiv:2112.10741}, 2021.

\bibitem{ref82}
R.~Rombach, A.~Blattmann, D.~Lorenz, P.~Esser, and B.~Ommer, ``High-resolution
  image synthesis with latent diffusion models,'' in \emph{Proc. IEEE Conf.
  Comput. Vis. Pattern Recognit.}, 2022, pp. 10\,684--10\,695.

\bibitem{ref83}
S.~Gu, D.~Chen, J.~Bao, F.~Wen, B.~Zhang, D.~Chen, L.~Yuan, and B.~Guo,
  ``Vector quantized diffusion model for text-to-image synthesis,'' in
  \emph{Proc. IEEE Conf. Comput. Vis. Pattern Recognit.}, 2022, pp.
  10\,696--10\,706.

\end{thebibliography}



%

\begin{IEEEbiography}[{\includegraphics[width=1in,height=1.25in,clip,keepaspectratio]{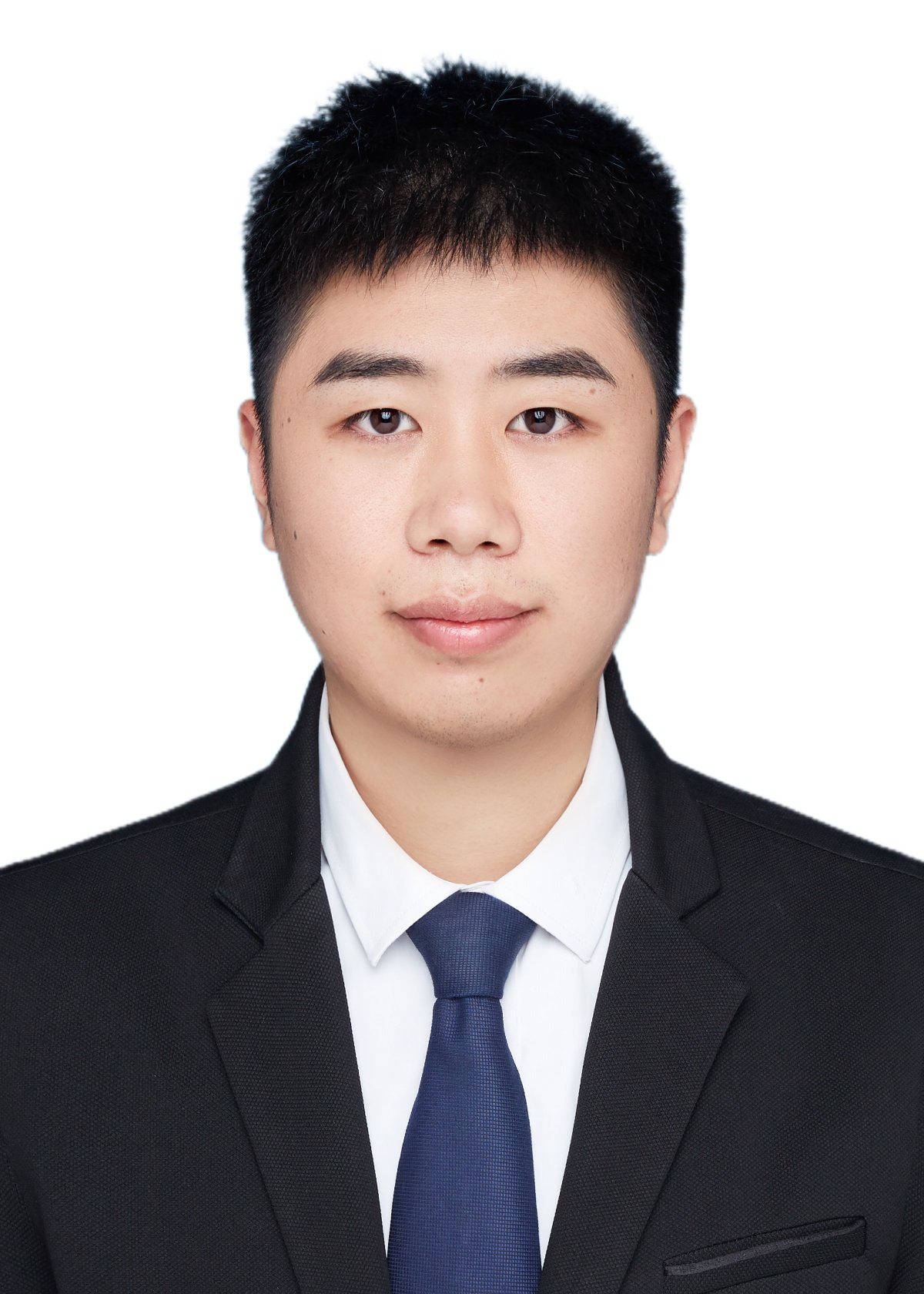}}]{Shuren Qi}
	received the B.A. and M.E. degrees from Liaoning Normal University, Dalian, China, in 2017 and 2020 respectively. He is currently pursuing the Ph.D. degree in computer science at Nanjing University of Aeronautics and Astronautics, Nanjing, China. He has published academic papers in top-tier venues including \emph{ACM Computing Surveys} and \emph{IEEE Transactions on Pattern Analysis and Machine Intelligence}. His research involves the general topics of invariance, robustness, and explainability in computer vision, with a focus on invariant representations, for closing today's trustworthiness gap in artificial intelligence, e.g., forensic and security of visual data.
\end{IEEEbiography}

\begin{IEEEbiography}[{\includegraphics[width=1in,height=1.25in,clip,keepaspectratio]{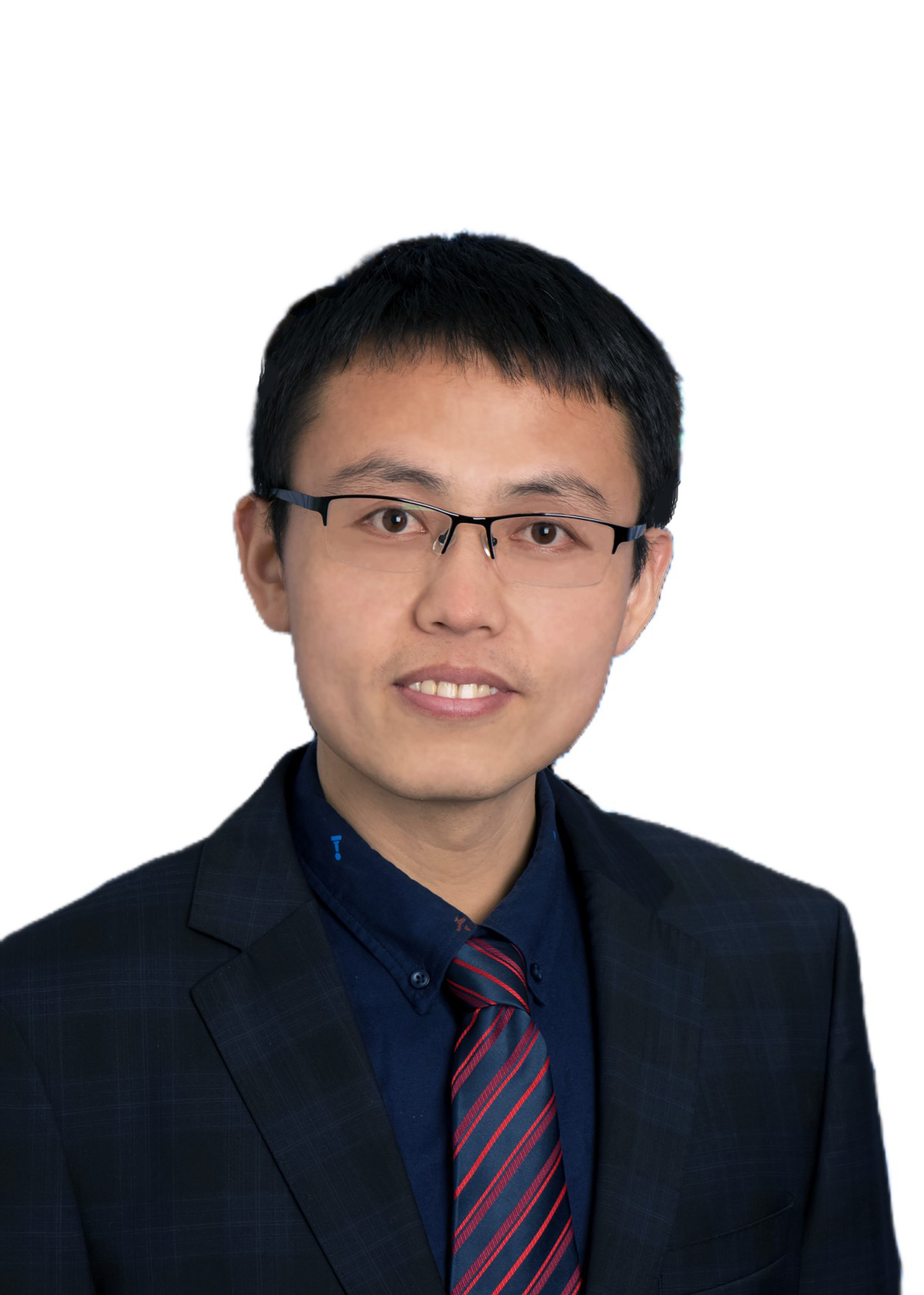}}]{Yushu Zhang}
	(Senior Member, IEEE) received the Ph.D. degree in computer science from Chongqing University, Chongqing, China, in 2014. He held various research positions with the City University of Hong Kong, Southwest University, University of Macau, and Deakin University. He is currently a Professor with the College of Computer Science and Technology, Nanjing University of Aeronautics and Astronautics, Nanjing, China. His research interests include multimedia processing and security, artificial intelligence, and blockchain. Dr. Zhang is an Associate Editor of \emph{Signal Processing} and \emph{Information Sciences}.
\end{IEEEbiography}
\vspace{-11 mm}

\begin{IEEEbiography}[{\includegraphics[width=1in,height=1.25in,clip,keepaspectratio]{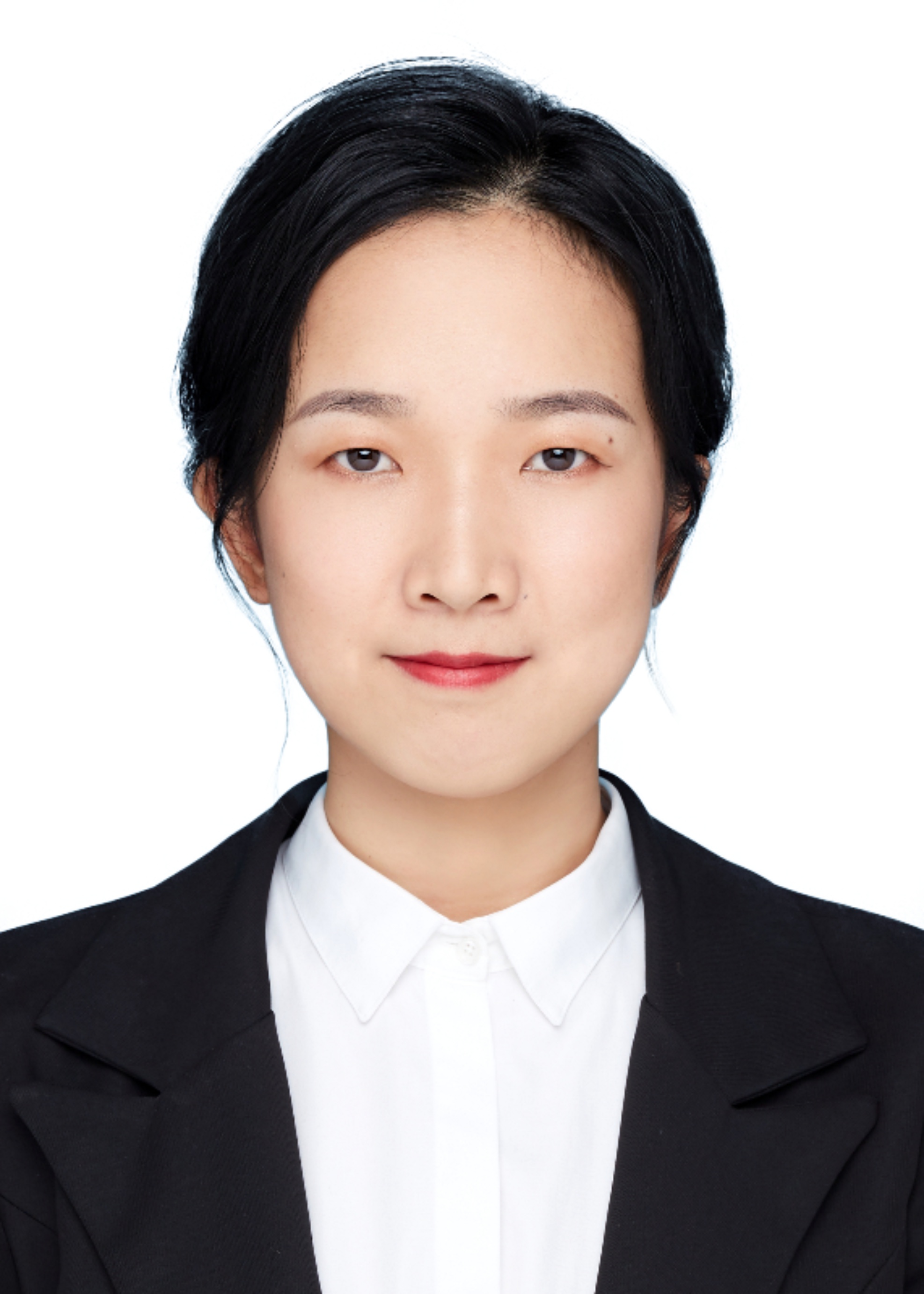}}]{Chao Wang}
	received the B.S. and M.S. degrees from Liaoning Normal University, Dalian, China, in 2017 and 2020 respectively. She is currently pursuing the Ph.D. degree in computer science at Nanjing University of Aeronautics and Astronautics, Nanjing, China. Her research interests include trustworthy artificial intelligence, adversarial learning, and media forensics.
\end{IEEEbiography}
\vspace{-11 mm}

\begin{IEEEbiography}[{\includegraphics[width=1in,height=1.25in,clip,keepaspectratio]{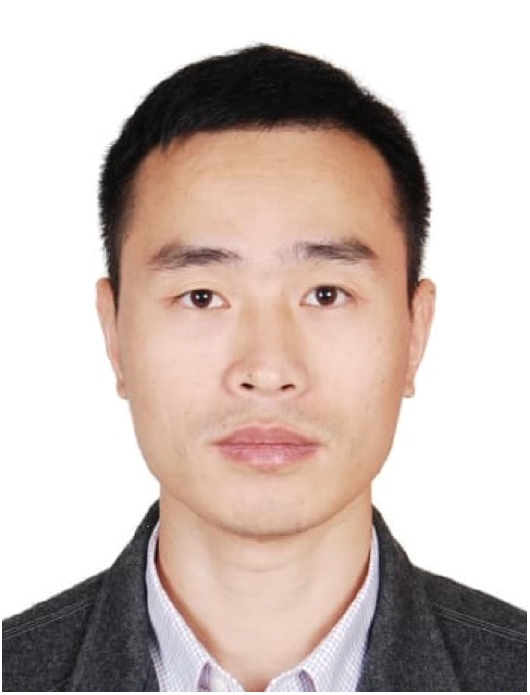}}]{Zhihua Xia}
	(Member, IEEE) received the Ph.D. degree in computer science from Hunan University, China, in 2011. He held various research positions with the Nanjing University of Information Science and Technology, New Jersey Institute of Technology, and Sungkyunkwan University. He is currently a Professor with the College of Cyber Security, Jinan University, China. His research interests include AI security, secure computation, and media forensics.
\end{IEEEbiography}
\vspace{-11 mm}

\begin{IEEEbiography}[{\includegraphics[width=1in,height=1.25in,clip,keepaspectratio]{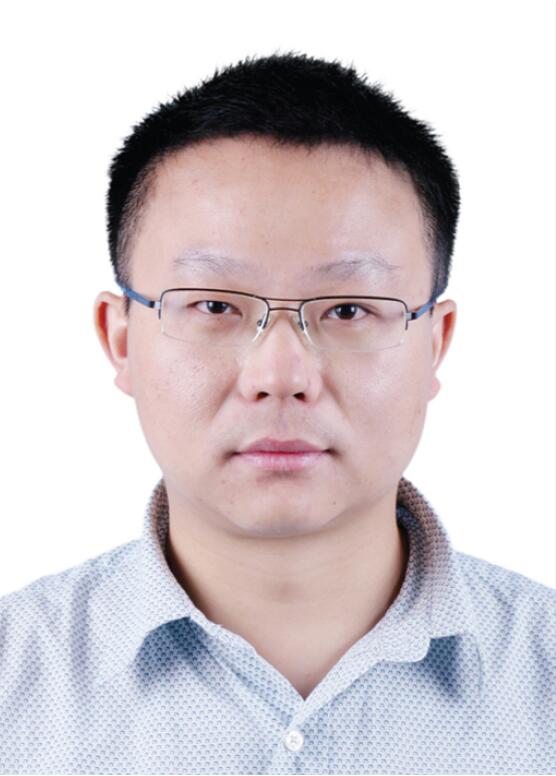}}]{Xiaochun Cao}
	(Senior Member, IEEE) received the B.E. and M.E. degrees in computer science from Beihang University, Beijing, China, in 1999 and 2002, respectively, and the Ph.D. degree in computer science from the University of Central Florida, Orlando, FL, USA, in 2006. After graduation, he spent about three years at ObjectVideo Inc., as a Research Scientist. From 2008 to 2012, he was a Professor at Tianjin University, Tianjin, China. Before joining Sun Yat-sen University, Shenzhen, China, he was a Professor at the Institute of Information Engineering, Chinese Academy of Sciences, Beijing, China. He is a Professor and the Dean with the School of Cyber Science and Technology, Shenzhen Campus of Sun Yat-sen University. He has published more than 200 journal and conference papers. Dr. Cao’s dissertation was nominated for the University Level Outstanding Dissertation Award. He was a recipient of the Piero Zamperoni Best Student Paper Award at the \emph{International Conference on Pattern Recognition}, in 2004 and 2010; the Excellent Young Scientists Fund and Distinguished Young Scholars Fund of National Natural Science Foundation of China, in 2014 and 2020. He was on the Editorial Boards of \emph{IEEE Transactions on Circuits and Systems for Video Technology} and \emph{IEEE Transactions on Multimedia}. He is on the Editorial Boards of \emph{IEEE Transactions on Pattern Analysis and Machine Intelligence} and \emph{IEEE Transactions on Image Processing}.
\end{IEEEbiography}
\vspace{-11 mm}

\begin{IEEEbiography}[{\includegraphics[width=1in,height=1.25in,clip,keepaspectratio]{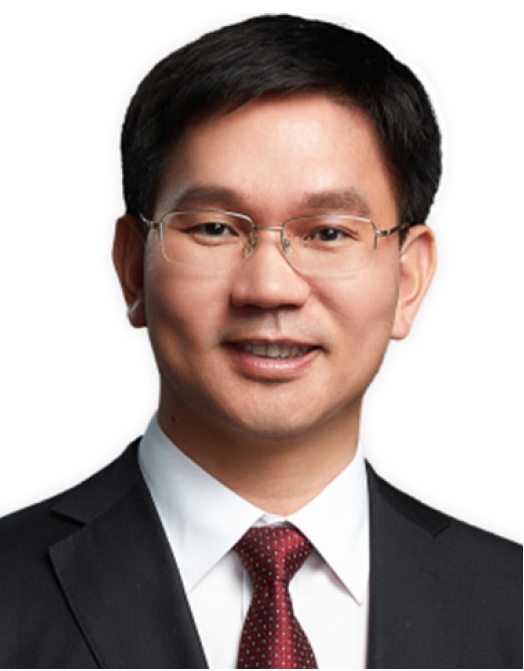}}]{Jian Weng}
	(Senior Member, IEEE) received the Ph.D. degree in computer science from Shanghai Jiao Tong University, Shanghai, China, in 2008. From 2008 to 2010, he was a Post-doctoral Researcher with the Singapore Management University, Singapore. He is currently a Professor and the Vice President of Jinan
	University, Guangzhou, China. He has published more than 100 journal and conference papers. Dr. Weng served as the PC co-chair or a PC member for more than 50 international conferences. He was a recipient of the Innovation Award from the Chinese Association for Cryptologic Research in 2015 and the Distinguished Young Scholars Fund of National Natural Science Foundation of China in 2018. He is on the Editorial Board of \emph{IEEE Transactions on Vehicular Technology}.
\end{IEEEbiography}




\end{document}